%% file: main.tex
\documentclass[10pt,twocolumn,letterpaper]{article}

\usepackage{cvpr}
\usepackage{times}
\usepackage{epsfig}
\usepackage{graphicx}
\usepackage{amsmath}
\usepackage{amssymb}
\usepackage{bbm}
\usepackage{subcaption}

\usepackage[pagebackref=true,breaklinks=true,colorlinks,bookmarks=false]{hyperref}

\cvprfinalcopy 

\def\cvprPaperID{0000} 

\usepackage{booktabs}
\usepackage[table]{xcolor}
\usepackage{multirow}

\usepackage[breaklinks=true,colorlinks,bookmarks=false]{hyperref}

\makeatletter
\usepackage{xspace}
\def\@onedot{\ifx\@let@token.\else.\null\fi\xspace}
\DeclareRobustCommand\onedot{\futurelet\@let@token\@onedot}

\newcommand{\figref}[1]{Fig\onedot~\ref{#1}}

\newcommand{\secref}[1]{Sec\onedot~\ref{#1}}
\newcommand{\tabref}[1]{Tab\onedot~\ref{#1}}

\def\eg{\emph{e.g}\onedot} 
\def\ie{\emph{i.e}\onedot} 
 
 \def\vs{\emph{vs}\onedot}
\def\wrt{w.r.t\onedot} 
\def\etal{\emph{et al}\onedot}

\ifcvprfinal\fi
\begin{document}

\title{DeeperLab: Single-Shot Image Parser}

\author{
  \begin{tabular}[t]{c}
Tien-Ju Yang$^{1}$, Maxwell D.~Collins$^2$, Yukun Zhu$^2$, Jyh-Jing Hwang$^{2,3}$, Ting Liu$^2$, \\
Xiao Zhang$^2$, Vivienne Sze$^1$, George Papandreou$^2$, Liang-Chieh Chen$^2$\\
MIT$^1$, Google Inc.$^2$, UC Berkeley $^3$\\
\end{tabular}
}

\maketitle


\input{0_abstract}


\input{1_introduction}

\input{2_related_work}

\input{3_methodology}

\input{4_experiment_results}

\input{5_conclusion}


{\small
\bibliographystyle{ieeetr}
\bibliography{__references}
}


\clearpage
\appendix
\input{6_appendix.tex}

\end{document}

%% file: 0_abstract.tex
\begin{abstract}
We present a single-shot, bottom-up approach for whole image parsing. Whole image parsing, also known as Panoptic Segmentation, generalizes the tasks of semantic segmentation for 'stuff' classes and instance segmentation for 'thing' classes, assigning both semantic and instance labels to every pixel in an image. Recent approaches to whole image parsing typically employ separate standalone modules for the constituent semantic and instance segmentation tasks and require multiple passes of inference. Instead, the proposed DeeperLab image parser performs whole image parsing with a significantly simpler, fully convolutional approach that jointly addresses the semantic and instance segmentation tasks in a single-shot manner, resulting in a streamlined system that better lends itself to fast processing. For quantitative evaluation, we use both the instance-based Panoptic Quality (PQ) metric and the proposed region-based Parsing Covering (PC) metric, which better captures the image parsing quality on 'stuff' classes and larger object instances. We report experimental results on the challenging Mapillary Vistas dataset, in which our single model achieves 31.95\% (val) / 31.6\% PQ (test) and 55.26\% PC (val) with 3 frames per second (fps) on GPU or near real-time speed (22.6 fps on GPU) with reduced accuracy.
\end{abstract}

%% file: 1_introduction.tex
\section{Introduction}
\label{sec:intro}

This paper addresses the problem of efficient whole image parsing (in short, image parsing)~\cite{tu2005image}, also known as \emph{Panoptic Segmentation}~\cite{kirillov2018panoptic}. Image parsing is a long-lasting unsolved problem in computer vision and a basic component of many applications, such as autonomous driving. The high difficulty lies in the fact that image parsing unifies two challenging tasks, semantic segmentation and instance segmentation. Semantic segmentation focuses on partitioning the whole image into multiple semantically meaningful regions, regardless of whether the semantic class is countable (a 'thing' class) or uncountable (a 'stuff' class). In contrast, instance segmentation only handles the region related to the 'thing' classes but requires telling different instances apart. As the combination of them, image parsing attempts to segment the whole image for both the 'thing' and 'stuff' classes and separate different 'thing' instances.

Although there have been a few works trying to solve the image parsing problem, efficiency is usually not taken into account. Efficiency is important for practical deployment. For example, Kirillov \etal~\cite{kirillov2018panoptic} report excellent parsing results, but the computational cost can be high due to the multiple passes of several complicated networks. The computation can also be more intense when high-resolution images are used as the input; \eg, the Mapillary Vistas dataset~\cite{neuhold2017mapillary} contains images with a resolution of up to $4000\times6000$.

\begin{figure}[!t]
    \centering
    \includegraphics[width=0.47\textwidth]{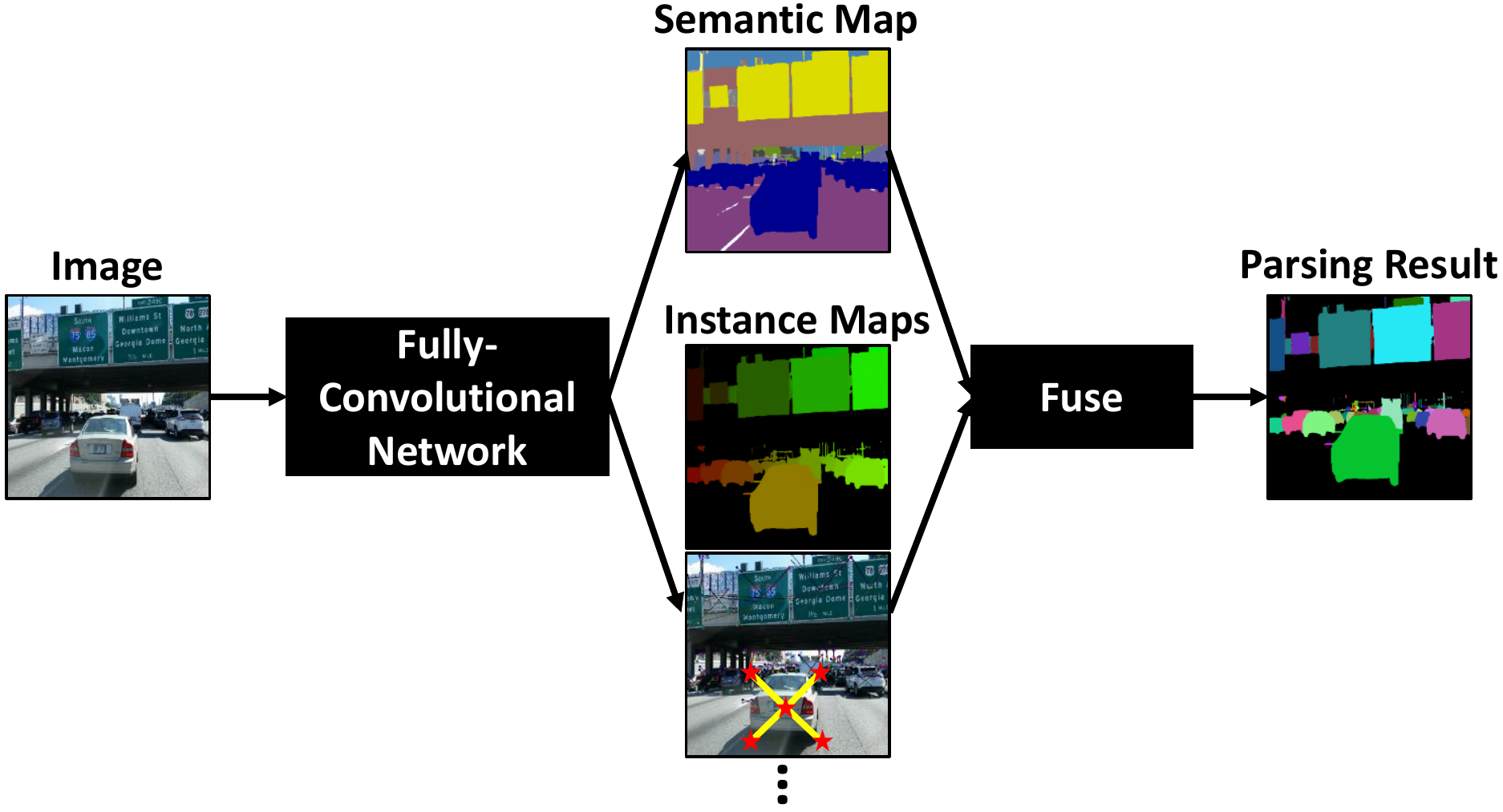}
    \caption{The proposed single-shot, bottom-up image parser, DeeperLab. The \emph{per-pixel} semantic and instance predictions are generated using a \emph{single} pass of a fully-convolutional network. These predictions are then fused into the final image parsing result by a fast algorithm.}
    \label{fig:simplified_network_architecture}
\end{figure}

In this work, we aim to design an image parser that achieves a good balance between accuracy and efficiency. We propose a single-shot, bottom-up image parser, called \emph{DeeperLab}. As shown in Fig.~\ref{fig:simplified_network_architecture}, DeeperLab generates the \emph{per-pixel} semantic and instance predictions using a \emph{single} pass of a fully-convolutional network. These predictions are then fused into the final image parsing result by a fast algorithm. The runtime of DeeperLab is nearly independent of the number of detected object instances, which makes DeeperLab favorable for image parsing of complex scenes.

For quantitative evaluation, we argue that the recently proposed instance-based Panoptic Quality (PQ) metric \cite{kirillov2018panoptic} often places disproportionate emphasis on small instance parsing, as well as on 'thing' over 'stuff' classes. To remedy these effects, we propose an alternative region-based \emph{Parsing Covering (PC)} metric\footnote{The code is available at http://deeperlab.mit.edu.}, which adapts the Covering metric~\cite{amfm_pami2011}, previously used for class-agnostics segmentation quality evaluation, to the task of image parsing. We report quantitative results with both PQ and PC metrics.

Our main contributions are summarized below.
\begin{itemize}
    \item We propose several neural network design strategies for efficient image parsers, especially reducing memory footprint for high-resolution inputs. These innovations include extensively applying depthwise separable convolution, using a shared decoder output with simple two-layer prediction heads, enlarging kernel sizes instead of making the network deeper, employing space-to-depth and depth-to-space rather than upsampling, and performing hard data mining. Detailed ablation studies are also provided to show the impact of these strategies in practice.
    \item We propose an efficient single-shot, bottom-up image parser, DeeperLab, based on the proposed design strategies. For example, on the Mapillary Vistas dataset, our Xception-71~\cite{chollet2016xception,dai2017coco,deeplabv3plus2018} based model achieves 31.95\% PQ (val) / 31.6\% (test) and 55.26\% PC (val) with 3 frames per second (fps) on GPU. Our novel \emph{Wider} version of the MobileNetV2~\cite{howard2017mobilenets} based model can achieve near real-time performance (22.61 fps on GPU) with reduced accuracy.
    \item We propose an alternative metric, Parsing Covering, to evaluate image parsing results from a region-based perspective.
\end{itemize}
We report results on additional datasets (Cityscapes, Pascal VOC 2012, and COCO) in the supplementary material.

%% file: 2_related_work.tex
\section{Related Work}
\label{sec:related_work}

{\bf Image parsing:} The task of image parsing refers to decomposing images into constituent visual patterns, such as textures and object instances. It unifies detection, segmentation, and recognition. Tu \etal \cite{tu2005image} present the first attempt for image parsing in a Bayesian framework. Since then, there have been several works aiming to jointly perform detection and segmentation for whole scene understanding with AND-OR graphs \cite{zhu2007stochastic,zhu2012recursive}, Exemplars \cite{malisiewicz2008recognition,tighe2013finding,isola2013scene}, or Conditional Random Fields \cite{rabinovich2007objects,gould2009region,heitz2009cascaded,ladicky2010and,yao2012describing,sun2014relating}. These early works evaluated image parsing results with separate metrics (\eg, one for object detection and one for semantic segmentation). There has been renewed interest in this task, also called Panoptic Segmentation, with the introduction of the unified instance-based Panoptic Quality (PQ) metric \cite{kirillov2018panoptic} into several benchmarks \cite{lin2014microsoft, neuhold2017mapillary}.

{\bf Semantic segmentation:} Most of the state-of-the-art semantic segmentation models are built upon fully convolutional neural networks (FCNs)~\cite{sermanet2013overfeat,long2014fully} and further improve the performance by incorporating different innovations. For example, it has been known that contextual information is essential for pixel labeling \cite{he2004multiscale,shotton2009textonboost,kohli2009robust,ladicky2009associative,gould2009decomposing,mostajabi2014feedforward,dai2015convolutional}. Following this idea, several works~\cite{farabet2013learning,eigen2015predicting,pinheiro2014recurrent,lin2015efficient,chen2015attention,chen2017deeplab} adopt image pyramids to encode contexts with different input sizes. Recently, PSPNet~\cite{zhao2017pyramid} proposes using spatial pyramid pooling~\cite{grauman2005pyramid,lazebnik2006beyond} at several grid scales (including image-level pooling~\cite{liu2015parsenet}), and DeepLab~\cite{chen2017deeplab,chen2017rethinking} proposes applying several parallel atrous convolutions~\cite{holschneider1989real,giusti2013fast,sermanet2013overfeat,papandreou2014untangling,chen2014semantic} with different rates (called Atrous Spatial Pyramid Pooling, or ASPP). By effectively utilizing the multi-scale contextual information, these models demonstrate promising accuracy on several segmentation benchmarks. Another effective way is the employment of the encoder-decoder structure~\cite{ronneberger2015u,badrinarayanan2015segnet,newell2016stacked}. Typically, the encoder-decoder networks~\cite{noh2015learning,ronneberger2015u,badrinarayanan2015segnet,lin2016refinenet,pohlen2016full,peng2017large,islamgated,wojna2017devil,fu2017stacked,deeplabv3plus2018,zhang2018exfuse,xiao2018unified} capture the context information in the encoder path and recover the object boundary in the decoder path. To maximize the accuracy on image parsing, the proposed DeeperLab utilizes most of these techniques, which are the FCN, ASPP and encoder-decoder structure.

{\bf Instance segmentation:} Current solutions for instance segmentation could be roughly categorized into top-down and bottom-up methods. The top-down approaches~\cite{dai2016instance,dai2017fully,dai2017deformable,he2017mask,chen2018masklab,liu2018path,peng2018megdet} obtain instance masks by refining the predicted boxes from state-of-the-art detectors \cite{ren2015faster,dai2016rfcn,cai2018cascade}. FCIS \cite{dai2017fully} employs the position-sensitive score maps \cite{dai2016instancesensitive}. Mask-RCNN \cite{he2017mask}, built on top of FPN \cite{lin2016feature}, attaches another segmentation branch to Faster-RCNN \cite{ren2015faster} and demonstrates outstanding performance. Additionally, some methods directly aim for mask proposals instead of bounding box proposals, including \cite{carreira2012cpmc,arbelaez2014multiscale,pinheiro2015learning,pinheiro2016learning,dai2016instancesensitive}. On the other hand, the bottom-up approaches~\cite{liang2015proposal,zhang2015monocular,uhrig2016pixel,zhang2016instance,bai2017deep,liu2017sgn,kirillov2017instancecut,newell2017associative,fathi2017semantic,de2017semantic,kendall2018multi,liu2018affinity} generally adopt a two-stage processing: pixel-level predictions produced by the segmentation module are clustered together to form the instance-level predictions. Recently, PersonLab \cite{papandreou2018personlab} predicts person keypoints and person instance segmentation, while DeNet \cite{tychsen2017denet} and CornerNet \cite{law2018cornernet} detect instances by predicting the corners of their bounding boxes. Our work is similar in the sense that we also produce keypoints for instance segmentation, which however is only part of our whole image parsing pipeline.

{\bf Evaluation metrics:} Semantic segmentation results can be evaluated by region-based metrics or contour-based metrics. Region-based metrics measure the proportion of correctly labelled pixels, including overall pixel accuracy~\cite{he2004multiscale,shotton2009textonboost}, mean class accuracy~\cite{shotton2009textonboost}, and mean IOU (intersection-over-union)~\cite{everingham2014pascal}. In contrast, contour-based metrics focus on the labeling precision around the segment boundaries. For example, \cite{kohli2009robust} measures the pixel accuracy or IOU within a Trimap in a narrow band around segment boundaries. Class-agnostic segmentation can be evaluated with the Covering metric~\cite{amfm_pami2011}. We refer the interested readers to~\cite{amfm_pami2011,csurka2013good} for an overview of the related literature.

Instance segmentation is usually formulated as mask detection~\cite{dai2016instance,dai2017fully,he2017mask}, considered as a refinement of bounding box detection. Thus, the task is typically measured with $AP^r$, which involves computing the intersection-over-union \wrt mask overlaps instead of box overlaps \cite{hariharan2014simultaneous}. The segmentation quality is evaluated by averaging $AP^r$ results at different mask overlap accuracy thresholds ranging from 0.5 to 0.95 \cite{lin2014microsoft,Cordts2016Cityscapes}. Another line of work~\cite{silberman2014instance,zhang2015monocular,zhang2016instance,ren2017end,liu2017sgn} adopts the region-based Covering metric to evaluate the instance segmentation results, which is applicable to methods that do not allow overlapped predictions.

Image parsing results can be evaluated with the instance-based Panoptic Quality (PQ) metric \cite{kirillov2018panoptic}, which treats all image regions with the same 'stuff' class as a single instance. An issue with the PQ metric is that all object instances are treated the same irrespective of their size, and thus the PQ metric may place disproportionate emphasis on small instances, as well as on 'things' over 'stuff' classes.

%% file: 3_methodology.tex
\section{Methodology}
\label{sec:methodology}

\begin{figure*}[!t]
    \centering
    \includegraphics[width=1.0\textwidth]{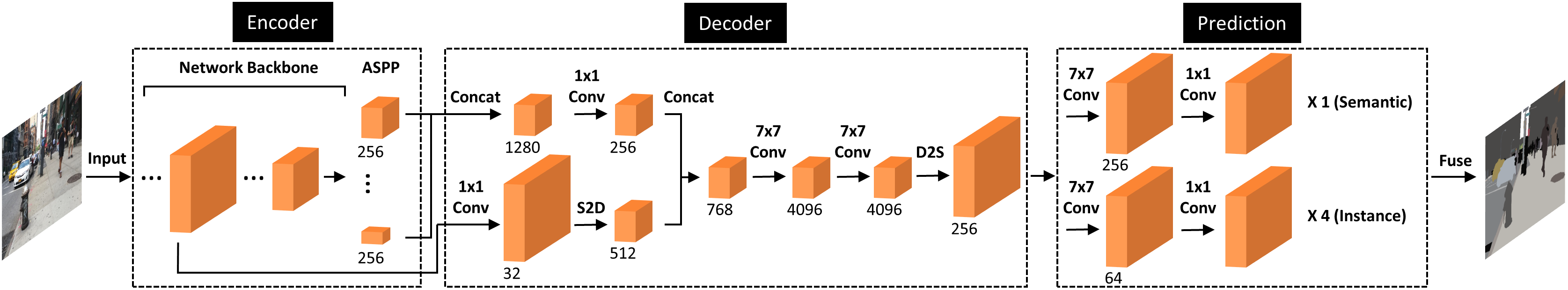}
    \caption{The proposed single-shot, bottom-up network architecture employs the encoder-decoder structure and produces per-pixel semantic and instance predictions. The number of channels of each feature map is specified in the figure.}
    \label{fig:network_architecture}
\end{figure*}

We propose an efficient single-shot, bottom-up neural network for image parsing, motivated by DeepLab~\cite{deeplabv3plus2018} and PersonLab~\cite{papandreou2018personlab}, which is illustrated in \figref{fig:network_architecture}. The proposed network adopts the encoder-decoder paradigm. For efficiency, the semantic segmentation and instance segmentation are generated from the \emph{shared decoder output} and then fused to produce the final image parsing result.

For image parsing, the network usually operates on high resolution inputs (\eg, $1441\times1441$ on resized Mapillary Vistas images in our experiments), which leads to high memory usage and latency. Below, we provide details of each component design regarding how we address this challenge and achieve a balance between accuracy and latency/memory footprint during both training and inference.

\subsection{Encoder}

We have experimented with two networks built on the efficient depthwise separable convolution~\cite{howard2017mobilenets}: the standard Xception-71~\cite{chollet2016xception,dai2017coco,deeplabv3plus2018} for higher accuracy, and a novel \emph{Wider} variant of MobileNetV2~\cite{mobilenetv22018} for faster inference.

Although standard MobileNetV2 performs well on the ImageNet image classification task with an input size of $224\times224$, it fails to capture long-range context information given its limited receptive field ($491\times491$ pixels) for the task of image parsing with high-resolution inputs. Stacking more $3\times3$ convolutions is a common practice to increase the receptive field, as is done in Xception-71. However, the extra layers introduce more feature maps, which dominate the memory usage. Considering the limited computation resources, we propose to replace all the $3\times3$ convolutions in MobileNetV2 with $5\times5$ convolutions. This approach efficiently increases the receptive field to $981\times981$ while maintaining the same amount of memory footprint for feature maps and only mildly increasing the computation cost. We refer to the resulting backbone as \emph{Wider MobileNetV2}.

Additionally, we augment the network backbone with the effective ASPP module (Atrous Spatial Pyramid Pooling)~\cite{chen2017deeplab,chen2017rethinking}. ASPP applies several parallel atrous convolutions with different rates to further increase the receptive field. The feature map at the encoder output has stride 16, \ie, its spatial resolution is equal to the input size downsampled by a factor of 16 across each spatial dimension.

\subsection{Decoder}

\begin{figure}[!t]
    \centering
    \includegraphics[width=0.3\textwidth]{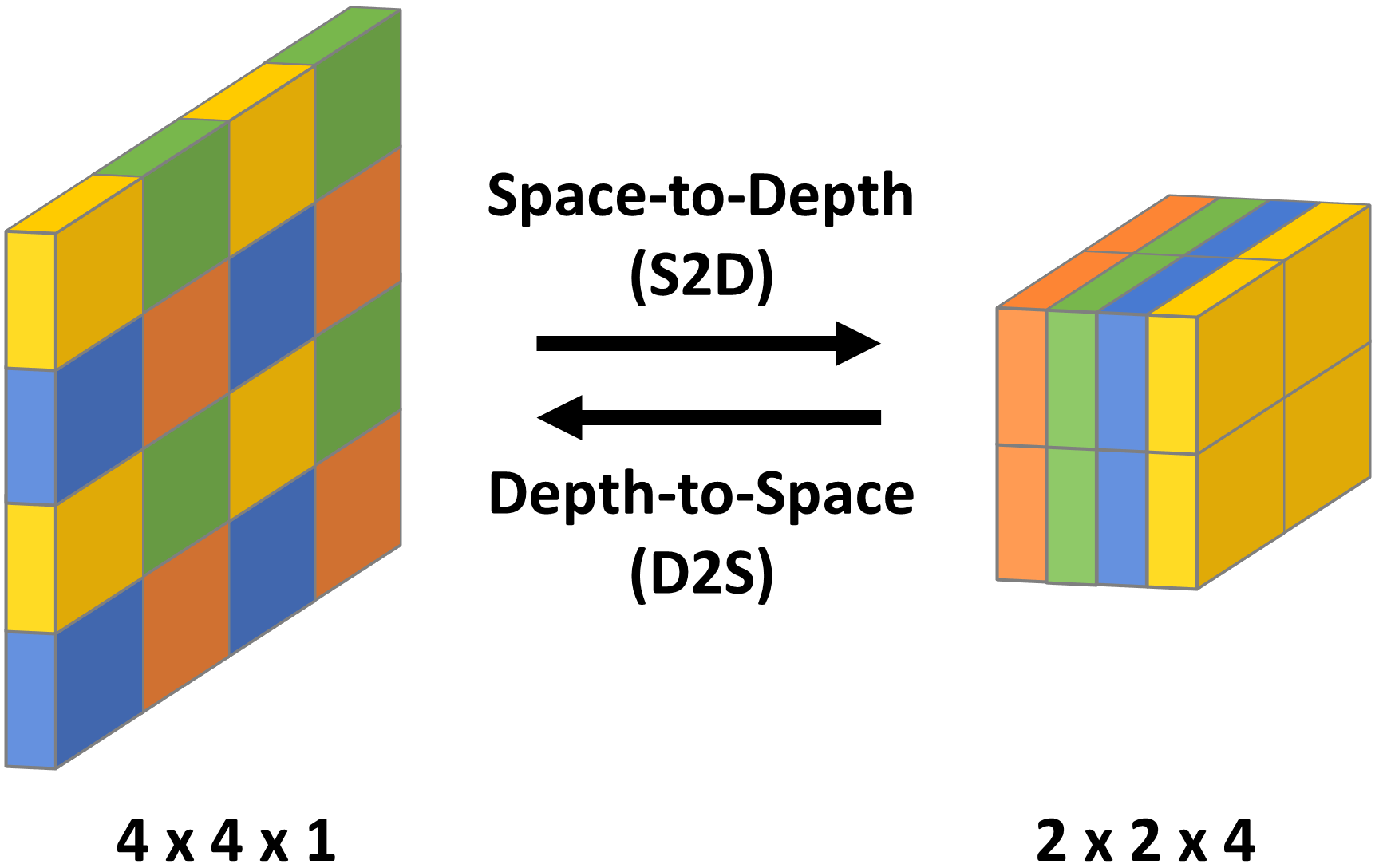}
    \caption{An example of the space-to-depth (S2D) and depth-to-space (D2S) operations. The S2D operation moves activations from the spatial dimension to the channel dimension and the D2S operation is the inverse.}
    \label{fig:s2d}
\end{figure}

The goal of the decoder module is to recover detailed object boundaries. Following DeepLabV3+~\cite{deeplabv3plus2018}, we adopt a simple design that combines the activations at the output of the encoder (with stride 16) with low-level feature maps from the network backbone (with stride 4). The number of channels of the concatenated ASPP outputs and the low-level feature map are first individually reduced by $1\times1$ convolution and then concatenated together. DeepLabV3+ bilinearly upsamples the reduced ASPP outputs before concatenation in order to account for the different spatial resolutions; however, the upsampling operation significantly increases the memory consumption. In this work, we apply the space-to-depth operation~\cite{shi2016real,sajjadi2018frame} (\figref{fig:s2d}) to the reduced low-level feature map, which keeps the memory usage of feature maps the same.

Similar to the encoder, the decoder uses two large kernel ($7\times7$) depthwise convolutions to further increase the receptive field. The resultant feature map has 4096 channels, which is then reduced by depth-to-space (\figref{fig:s2d}, reverse operation of space-to-depth), yielding a feature map with 256 channels and stride 4, which are used as the input for the image parsing prediction heads.

\subsection{Image Parsing Prediction Heads}

The proposed network contains five prediction heads, each of which is directly attached to the shared decoder output and consists of two convolution layers with kernel sizes of $7\times7$ and $1\times1$ respectively. One head (with 256 filters for the first $7\times7$ layer) is specific for semantic segmentation, while the other four (each with 64 filters for the first $7\times7$ layer) are used for class-agnostic instance segmentation.

\subsubsection{Semantic Segmentation Head}
\label{subsubsec:semantic_segmentation_predicton}

The semantic segmentation prediction is trained to minimize the bootstrapped cross-entropy loss \cite{wu2016bridging, bulo2017loss, pohlen2016full}, in which we sort the pixels based on the cross-entropy loss and we only backpropagate the errors in the top-K positions (hard example mining). We set $K = 0.15 \cdot N$, where $N$ is the total number of pixels in the image. Moreover, we weigh the pixel loss based on instance sizes, putting more emphasis on small instances. Specifically, our proposed weighted bootstrapped cross-entropy loss is defined by:

\begin{align} 
  \label{eq:semantic_loss}
  \ell = & - \frac{1}{K} \sum_{i=1}^N w_i \cdot \mathbbm{1}[p_{i,y_i} < t_K] \cdot \log{p_{i,y_i}},
\end{align} 
where $y_i$ is the target class label for pixel $i$, $p_{i,j}$ is the predicted posterior probability for pixel $i$ and class $j$, and $\mathbbm{1}[x] = 1$ if $x$ is true and 0 otherwise. The threshold $t_K$ is set in a way that only the pixels with top-K highest losses are selected. We set the weight $w_i = 3$ for pixels that belong to instances with an area smaller than $64 \times 64$ and $w_i = 1$ everywhere else. By doing so, the network is trained to focus on both hard pixels and small instances.

\subsubsection{Instance Segmentation Heads}
\label{subsubsec:instance_segmentation_prediction}

Similar to \cite{papandreou2018personlab, tychsen2017denet, law2018cornernet}, we adopt a \emph{keypoint-based} representation for object instances. In particular, we consider the four bounding box corners and the center of mass as our $P = 5$ object keypoints.

Following PersonLab~\cite{papandreou2018personlab}, we define four prediction heads, which are used for instance segmentation: a keypoint heatmap as well as long-range, short-range, and middle-range offset maps. Those predictions focus on predicting different relations between each pixel and the keypoints of its corresponding instance, which we fuse to form the class-agnostic instance segmentation as detailed in \secref{sec:fuse_instance}.

\begin{figure}
     \centering
     \begin{subfigure}[b]{0.1\textwidth}
         \centering
         \includegraphics[width=\textwidth]{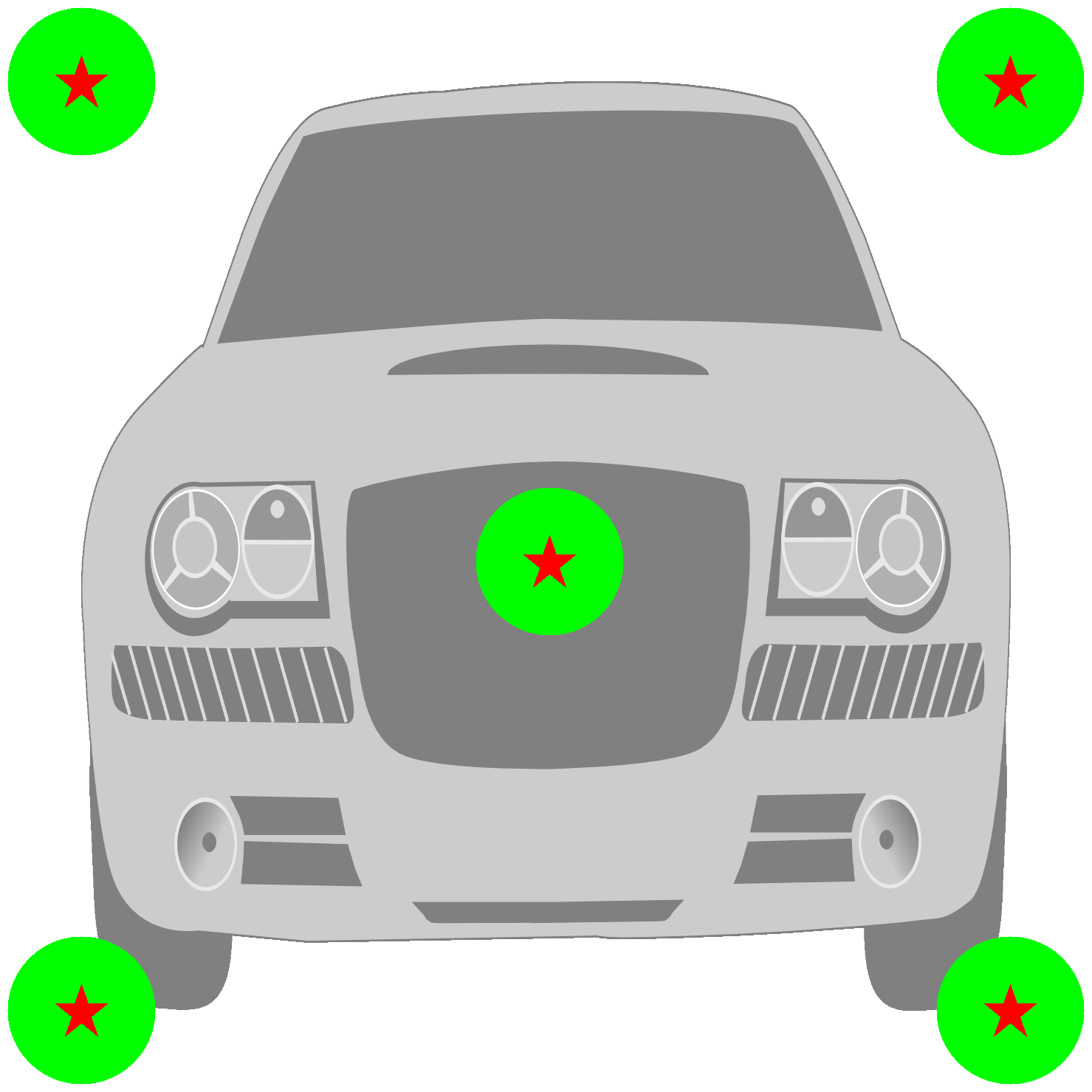}
         \caption{}
         \label{fig:heatmap}
     \end{subfigure}
     \hfill
     \begin{subfigure}[b]{0.1\textwidth}
         \centering
         \includegraphics[width=\textwidth]{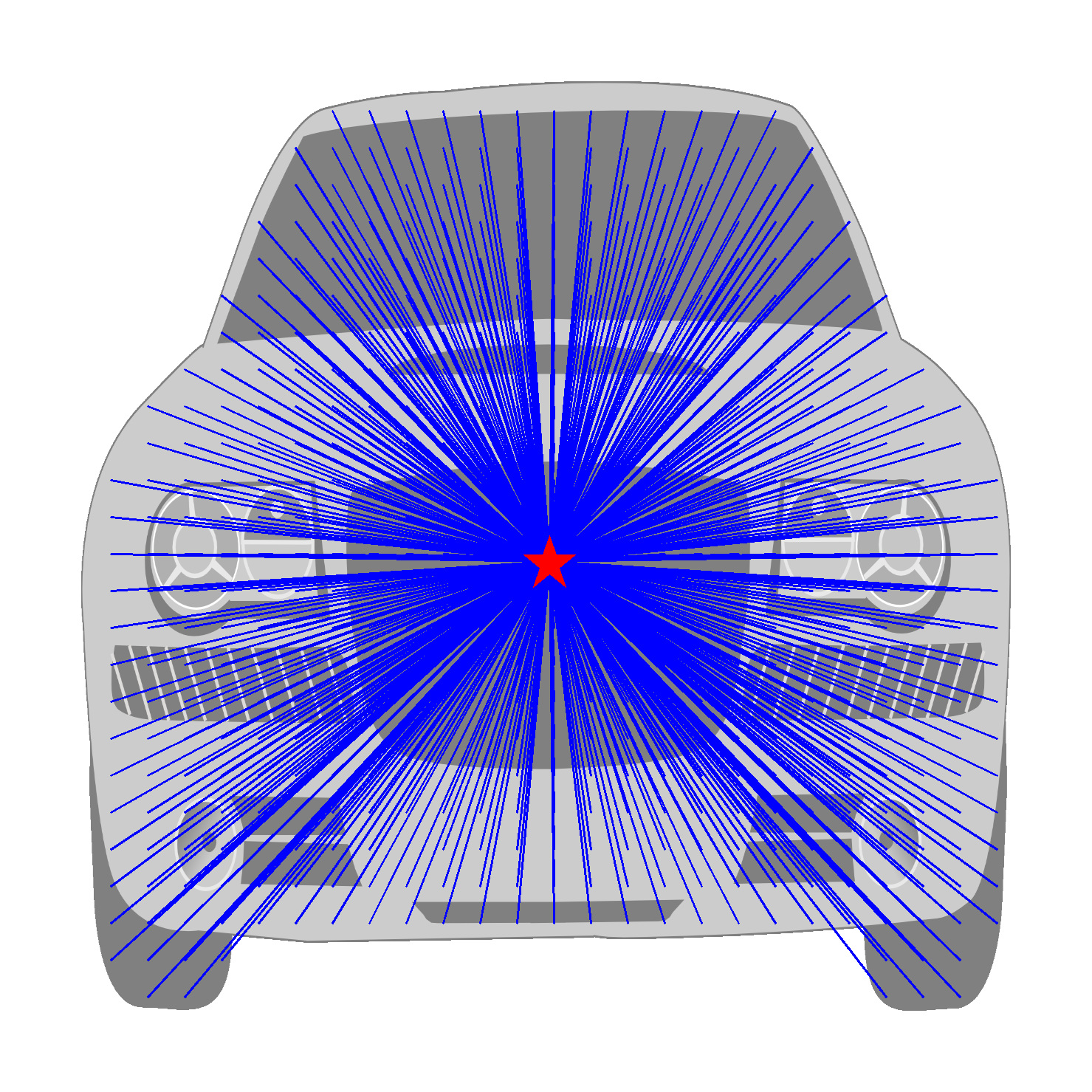}
         \caption{}
         \label{fig:long_range_offset_map}
     \end{subfigure}
     \hfill
     \begin{subfigure}[b]{0.1\textwidth}
         \centering
         \includegraphics[width=\textwidth]{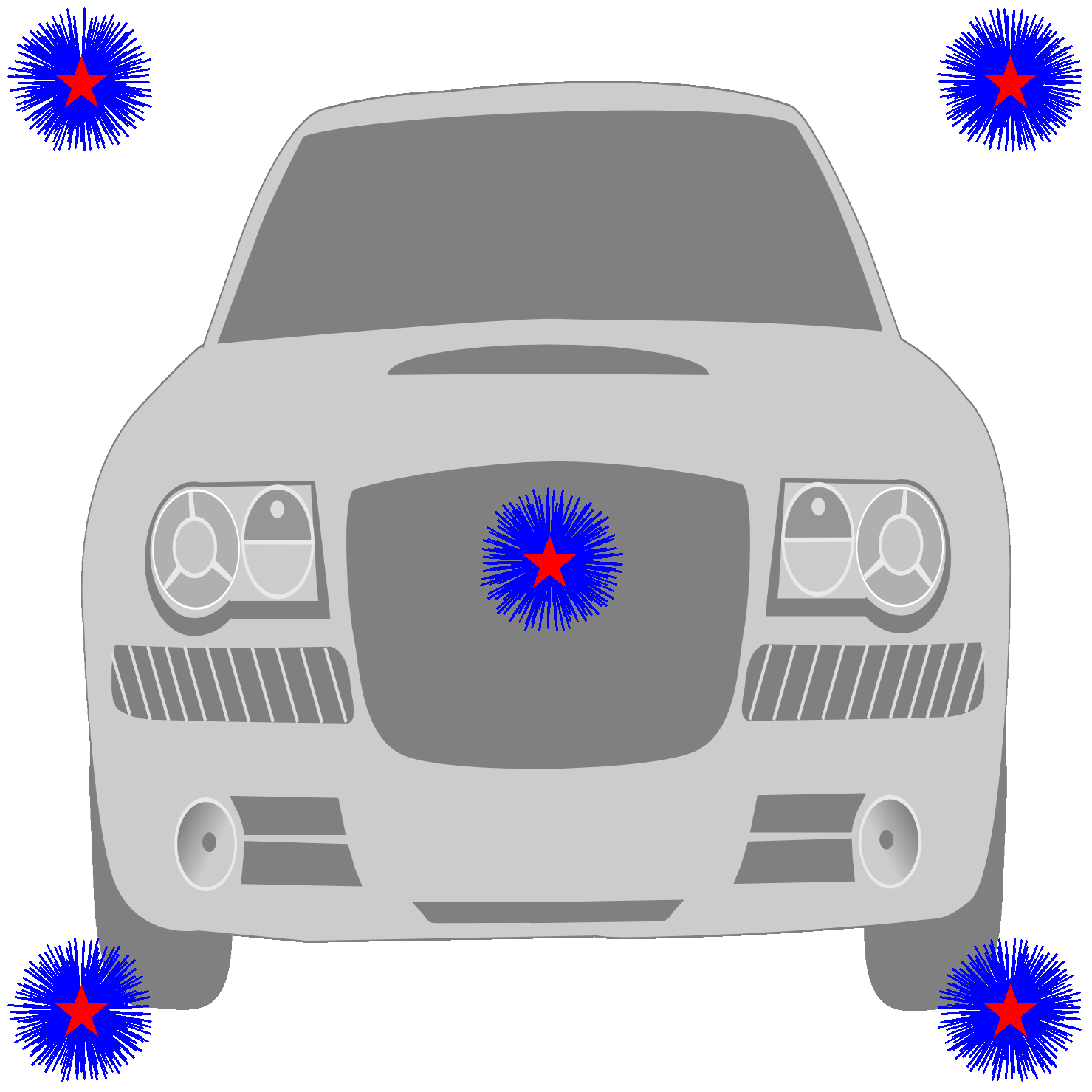}
         \caption{}
         \label{fig:short_range_offset_map}
     \end{subfigure}
     \hfill
     \begin{subfigure}[b]{0.1\textwidth}
         \centering
         \includegraphics[width=\textwidth]{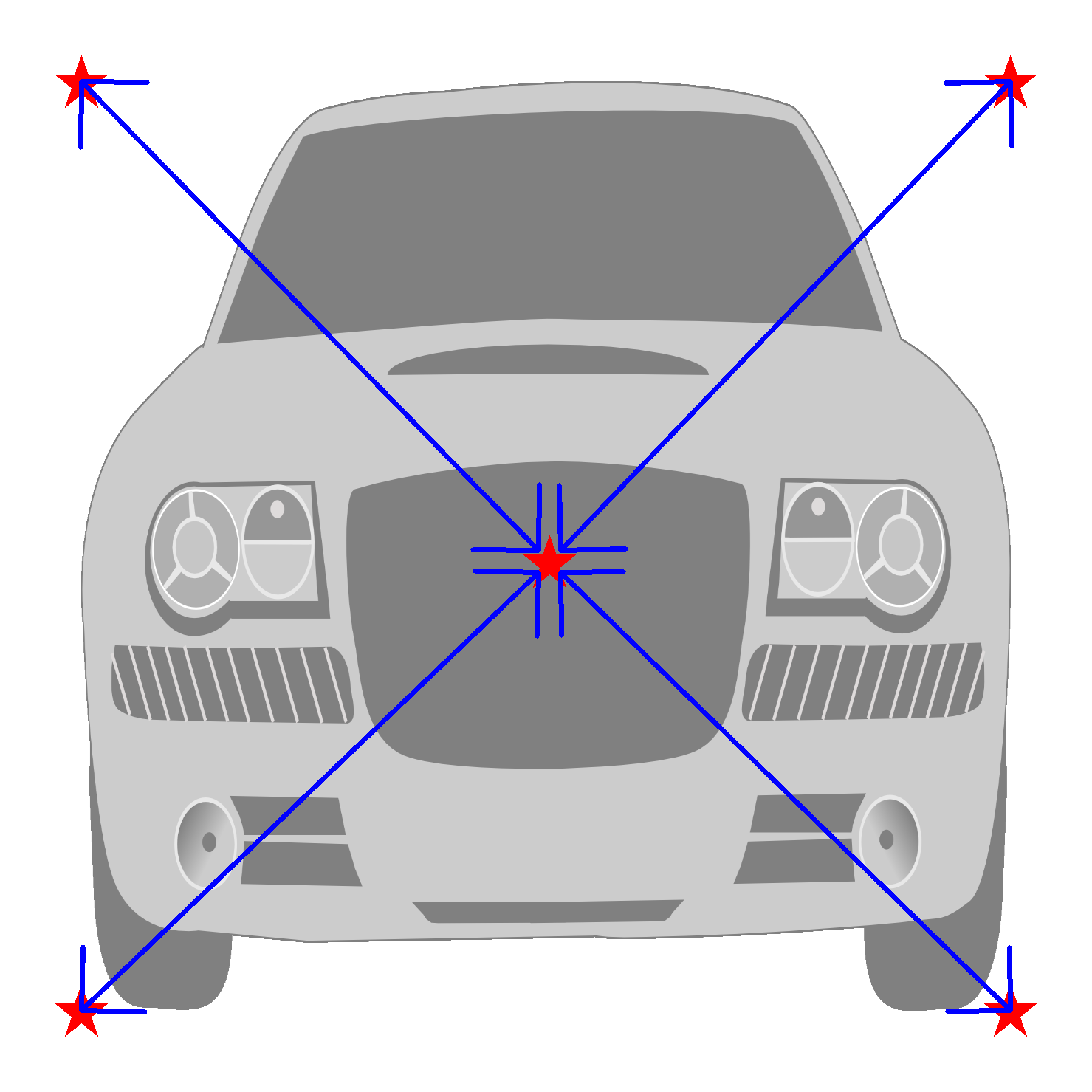}
         \caption{}
         \label{fig:middle_range_offset_map}
     \end{subfigure}
        \caption{Four prediction maps generated by our instance-related heads: (a) keypoint heatmap, (b) long-range offset, (c) short-range offset, and (d) middle-range offset. The red stars denote the keypoints, the green disk denotes the target for keypoint prediction, and the blue lines/arrows denote the offsets from the current pixel to the target keypoint.}
        \label{fig:instance_predictions}
\end{figure}

\textbf{The keypoint heatmap} (\figref{fig:heatmap}) predicts whether a pixel is within a disk of radius $R$ pixels centered in the corresponding keypoint. The target activation is equal to 1 in the interior of the disks and 0 elsewhere. We use the same disk radius $R=25$ regardless of the size of an instance, so that the network pays the same attention to both large and small instances. The predicted keypoint heatmap contains $P$ channels, one for each keypoint. We penalize prediction errors by the standard sigmoid cross entropy loss.

\textbf{The long-range offset map} (\figref{fig:long_range_offset_map}) predicts the position offset from a pixel to all the corresponding keypoints, encoding the \textit{long-range} information for each pixel. The predicted long-range offset map has $2P$ channels, where every two channels predict the offset in the horizontal and vertical directions for each keypoint. We employ $L_1$ loss for long-range offset prediction, which is only activated at pixels belonging to object instances.

\textbf{The short-range offset map} (\figref{fig:short_range_offset_map}) is similar to the long-range offset map except that it only focuses on pixels within the disk of radius $R = 25$ pixels around the keypoints, \ie, the pixels having a value of one in the target heatmap (the green disk in \figref{fig:heatmap}). The short range offset map also has $2P$ channels and are used to improve keypoint localization. We employ $L_1$ loss, which is only activated at the interior of the disks.

\textbf{The middle-range offset map} (\figref{fig:middle_range_offset_map}) predicts the offset among keypoint pairs, defined in a \emph{directed keypoint relation graph (DKRG)}. This map is used to group keypoints that belong to the same instance (\ie, instance detection via keypoints). As shown in \figref{fig:middle_range_offset_map}, we adopt the star graph \cite{veksler2008star}, where the mass center is bi-directionally connected to the other four box corners. The predicted middle-range offset map has $2E$ channels, where $E$ is the number of directed edges in the DKRG ($E=8$ in the star graph). Similarly, we use two channels for each edge to predict the horizontal and vertical offsets and employ $L_1$ loss during training, which is only activated at the interior of the disks.

\subsection{Prediction Fusion}
\label{sec:fusion}
We first explain how to merge the four predictions (keypoint heatmap, long-range, short-range, and middle-range offset maps) into a single class-agnostic instance segmentation map. Given the predicted semantic and instance segmentation maps, the final fusion step assigns both semantic and instance labels to every pixel in the image.

\subsubsection{Instance Prediction}
\label{sec:fuse_instance}
We generate the instance segmentation map from the four instance-related prediction maps similarly to PersonLab~\cite{papandreou2018personlab}. We will highlight the main steps and differences in the following paragraphs.

\textbf{Recursive offset refinement:} We observe that the predictions that are closer to the corresponding keypoints are more accurate. Therefore, we recursively refine the offset maps by itself and/or each other as in PersonLab~\cite{papandreou2018personlab}.

\textbf{Keypoint localization:} For each keypoint, we perform Hough-voting on the short-range offset map and use the corresponding value in the keypoint heatmap (after sigmoid activation) as the voting weight to generate the short-range score map. We propose to also perform Hough-voting on the long-range offset map (using a weight equal to one for every vote) to generate the long-range score map. These two score maps are merged into one by taking per-pixel weighted sum. We then localize the keypoints by finding the local maxima in the resultant fused score map. Finally, we use the \emph{Expected-OKS} score~\cite{papandreou2018personlab} to rescore all keypoints.

\textbf{Instance detection: } We cluster the keypoints to detect instances by using a fast greedy algorithm. All the keypoints are first pushed into a priority queue and popped one at a time. If the popped keypoint is in the proximity of the corresponding keypoint of an already detected instance, we reject it and continue the process. Otherwise, we follow the predicted middle-range offsets to identify the positions of the remaining four keypoints, thus forming a newly detected instance. The confidence score of the detected instance is defined as the average of its keypoint scores. After all the instances are detected, we use bounding box non-maximum suppression to remove overlapping instances.

\textbf{Assignment of pixels to instances:} Finally, given the detected instances, we assign an instance label to each pixel by using the predicted long-range offset map, which encodes the pixel-wise offset to the keypoints. Specifically, we assign each pixel to the detected instance whose keypoints have the smallest $L_2$-distance to the pixel's predicted keypoints (\ie, its image location plus its predicted long-range offset).

\subsubsection{Semantic and Instance Prediction Fusion}
\label{sec:fuse_semantic_instance}
We opt for a simple merging method without any other post-processing, such as removal of small isolated regions in the segmentation maps. In particular, we start from the predicted semantic segmentation by considering `stuff' (\eg, sky) and `thing' (\eg, person) classes separately. Pixels predicted to have a `stuff` class are assigned with a single unique instance label. For the other pixels, their instance labels are determined from the instance segmentation result while their semantic labels are resolved by the majority vote of the corresponding predicted semantic labels.

\subsection{Evaluation Metrics}

Herein, we briefly review the Panoptic Quality (PQ) metric~\cite{kirillov2018panoptic} and propose the Parsing Covering (PC) metric, extended from the existing Covering metric \cite{amfm_pami2011}.

Given a groundtruth segmentation $S$ and a predicted segmentation $S'$, PQ is defined as follows:

\begin{align}
    \label{eq:pq}
    PQ = & \frac{\sum_{(R, R')\in TP}IOU(R, R')}{|TP| + \frac{1}{2}|FP| + \frac{1}{2}|FN|},
\end{align}
where $R$ and $R'$ are groundtruth regions and predicted regions respectively, and $|TP|$, $|FP|$, and $|FN|$ are the number of true positives, false postives, and false negatives. The matching is determined by a threshold of 0.5 Intersection-Over-Union (IOU).

PQ treats all regions of the same `stuff` class as one instance, and the size of instances is not considered. For example, instances with $10\times10$ pixels contribute equally to the metric as instances with $1000\times1000$ pixels. Therefore, PQ is sensitive to false positives with small regions and some heuristics could improve the performance, such as removing those small regions (as also pointed out in the open-sourced evaluation code from \cite{kirillov2018panoptic}). Thus, we argue that PQ is suitable in applications where one cares equally for the parsing quality of instances irrespective of their sizes.

There are applications where one pays more attention to large objects, \eg, portrait segmentation (where large people should be segmented perfectly) or autonomous driving (where nearby objects are more important than far away ones). Motivated by this, we propose to also evaluate the quality of image parsing results by extending the existing Covering metric \cite{amfm_pami2011}, which accounts for instance sizes. Specifically, our proposed metric, Parsing Covering (PC), is defined as follows:
\begin{align} 
  \label{eq:pc_1}
  Cov_i = & \frac{1}{N_i}\sum_{R\in S_i}\left | R \right | \cdot \max_{R'\in S_i'}IOU(R,R'), \\
  \label{eq:pc_2}
  N_i = & \sum_{R\in S_i}\left | R \right |, \\
  \label{eq:pc_3}
  PC = & \frac{1}{C}\sum_{i=1}^{C}Cov_i,
\end{align} 
where $S_i$ and $S_i'$ are the groundtruth segmentation and predicted segmentation for the $i$-th semantic class respectively, and $N_i$ is the total number of pixels of groundtruth regions from $S_i$. The Covering for class $i$, $Cov_i$, is computed in the same way as the original Covering metric except that only groundtruth regions from $S_i$ and predicted regions from $S_i'$ are considered. PC is then obtained by computing the average of $Cov_i$ over $C$ semantic classes. We plan to open-source our implementation of the PC metric to facilitate its adoption by other researchers.

We note that Covering has been used in several instance segmentation works \cite{silberman2014instance,zhang2015monocular,zhang2016instance,ren2017end,liu2017sgn}. The proposed PC is a simple extension of the Covering to evaluate image parsing results. It was pointed out in \cite{zhang2016instance} that Covering does not penalize the false positives. This is because, in \cite{zhang2016instance}, the Covering for the background class is not evaluated, which absorbs other classes' false positives. In the case of image parsing, this will not happen since \textit{all} the classes and every pixel will be taken into account.

Another notable difference between PQ and the proposed PC is that there is no matching involved in PC and hence no matching threshold. As an attempt to treat equally `thing` and `stuff`, the segmentation of `stuff` classes still receives partial PC score if the segmentation is only partially correct. For example, if one out of three equally-sized trees is perfectly segmented, the model will get the same partial score by using PC regardless of considering `tree` as `stuff` or `thing`.

%% file: 4_experiment_results.tex
\section{Experimental Results}
\label{sec:experiments}

We demonstrate the effectiveness and efficiency of DeeperLab and present ablation studies on the Mapillary Vistas~\cite{neuhold2017mapillary}. This dataset contains 66 semantic classes in a variety of traffic-related images, whose sizes range from $1024\times768$ to higher than $4000\times6000$. We report both Panoptic Quality (PQ) \cite{kirillov2018panoptic} and the proposed Parsing Covering (PC) for accuracy, and speed on a desktop CPU and GPU\footnote{CPU: Intel(R) Xeon(R) CPU E5-1650 v3 @ 3.50GHz, GPU: Tesla V100-SXM2}. We report results on other datasets (Cityscapes, Pascal VOC 2012, and COCO) in the supplementary material.

All the models are trained end-to-end without piecewise pretraining of each component except that the backbone is pretrained on ImageNet-1K~\cite{ILSVRC15}. The training configuration is the same as that in~\cite{chen2017rethinking}. In short, we employ the same learning rate schedule (\ie, ``poly'' policy~\cite{liu2015parsenet} with an initial learning rate of $0.01$), fine-tune batch normalization~\cite{ioffe2015batch} parameters for all layers, and use random scale data augmentation during training. The training batch sizes are $28$ and $16$ when employing MobileNetV2 (MNV2) ~\cite{mobilenetv22018} and Xception-71~\cite{chollet2016xception, dai2017coco, deeplabv3plus2018} as the network backbone, respectively. Similar to~\cite{zhao2017pyramid, neuhold2017mapillary}, we resize the images to 1441 pixels at the longest side to handle the large input variations on Mapillary Vistas and randomly crop $721\times721$ patches during training.

Our numbers are reported with \emph{single-scale} inference. Moreover, we do not employ any heuristic post-processing such as small region removal or assigning VOID label to low confidence predictions.

\subsection{Performance on Mapillary Vistas}

DeeperLab aims to achieve a balance between accuracy and speed, which facilitates the deployment of image parsing. In this section, we will analyze both of the accuracy and speed of the proposed DeeperLab\footnote{To the best of our knowledge, there is no 1) peer-reviewed works that release the models or report the latency numbers with 2) single-model, 3) single-scale settings on Mapillary Vistas at the time of the preparation of this work.}.

\begin{figure*}[!t]
    \centering
    \includegraphics[width=0.97\textwidth,keepaspectratio]{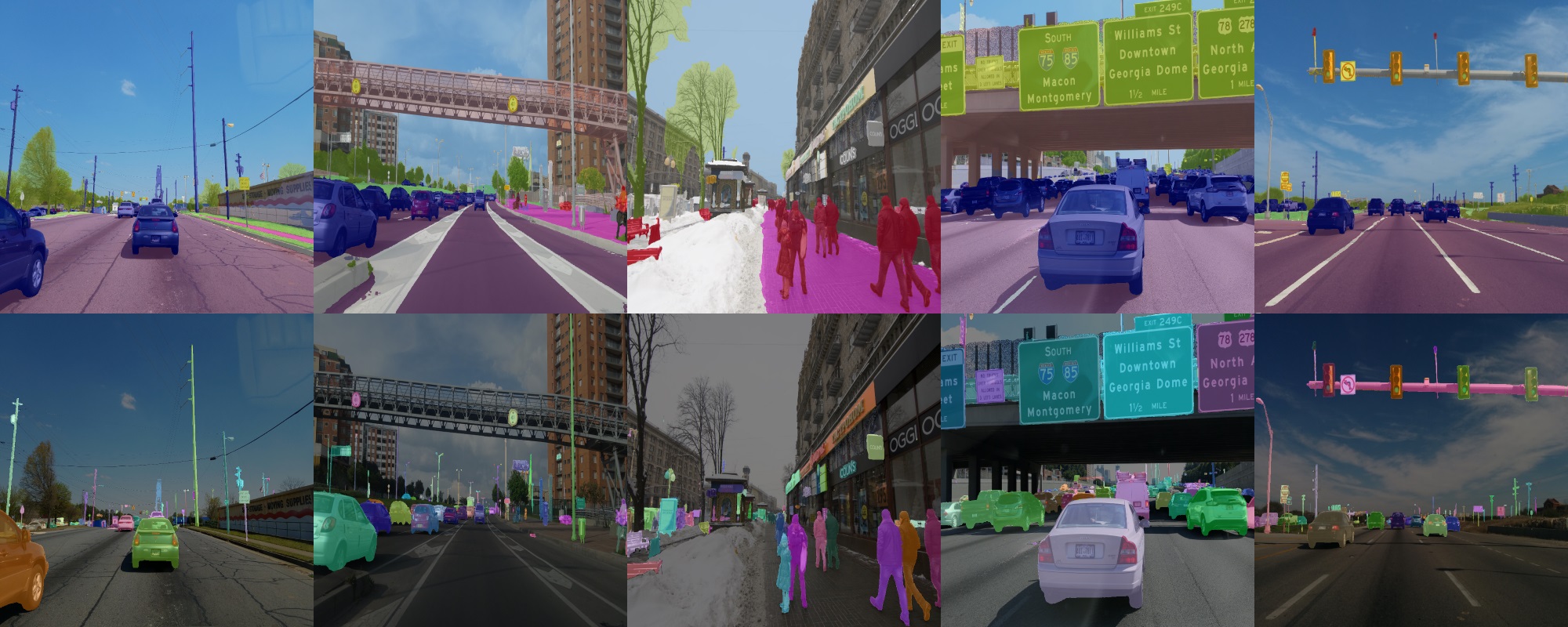}
    \caption{A few image parsing results on the Mapillary Vistas validation set with proposed DeeperLab based on Xception-71. The first row is the predicted semantic segmentation and the second row is the predicted instance segmentation. Note that our model does not generate any VOID labels.}
    \label{fig:xception_visualization_mapillary}
\end{figure*}

{\bf Validation set performance:} We summarize the accuracy and speed of DeeperLab on the validation set in~\tabref{tab:mapillary_val}, where the networks are trained longer than those in the ablation study, 500K \vs 200K iterations, respectively. Our Xception-71 based model\footnote{Our Xception-71 based model attains mIOU 55.30\% for the semantic segmentation task, outperforming 53.12\% in \cite{bulo2017place}.} attains 31.95\% PQ and 55.26\% PC, while the Wider MobileNetV2 based model achieves 25.20\% PQ and 49.80\% PC with faster inference (6.19 \vs 3.09 fps on GPU). We have also experimented with an even faster \emph{Light Wider MobileNetV2} variant, which employs a simpler decoder with $3\times3$ kernels and fewer filters (128 instead of 256 filters). The speed increases to 9.37 fps on GPU with a small accuracy drop. Additionally, if we downsample the input by 2, the Light Wider MobileNetV2 model can reach near real-time speed (22.61 fps on GPU). Note that we have not used any other tricks, such as folding batch norm or quantizing the models, to further speed up inference. Moreover, the extra step of fusing the semantic and instance segmentation is fast and mostly determined by the input resolution (145 ms for $1441\times1441$ image and 45 ms for $721\times721$ image with an unoptimized CPU implementation). \figref{fig:xception_visualization_mapillary} shows the qualitative results.

\begin{table}[!t]
  \centering
  \scalebox{0.58}{
  \begin{tabular}{c c | c c c c c}
    \toprule[0.2em]
    Method & Input Size & PQ (\%) & PC (\%) & fps (CPU) & fps (GPU) & Merge (ms)\\
    \toprule[0.2em]
    Light Wider MNV2 & $721\times721$ & 17.59 & 43.43 & 0.77 & 22.61 & 45 \\
    Light Wider MNV2 & $1441\times1441$ & 22.36 & 47.52 & 0.18 & 9.37 & 145 \\
    Wider MNV2 & $1441\times1441$ & 25.20 & 49.80 & 0.09 & 6.19 & 145 \\
    Xception-71 & $1441\times1441$ & 31.95 & 55.26 & 0.06 & 3.09 & 145 \\
    \bottomrule[0.1em]
  \end{tabular}
  }
  \caption{DeeperLab performance on the Mapillary Vistas validation set. Xception-71 based model attains higher accuracy while Wider MobileNetV2 (Wider MNV2) based model achieves faster inference. The model can be sped up by simplifying the decoder structure (Light Wider MNV2) with a small accuracy drop. With downsampled inputs, Light Wider MNV2 can reach near real-time speed.}
  \label{tab:mapillary_val}
\end{table}

{\bf Test set performance:} Our test set result is summarized in \tabref{tab:mapillary_test}, where only PQ is provided by the test server.

\begin{table}[!t]
  \centering
  \scalebox{0.61}{
  \begin{tabular}{c | c c c | c c c | c c c}
    \toprule[0.2em]
    Method & PQ & SQ & RQ & PQ$^{\text{Th}}$ & SQ$^{\text{Th}}$ & RQ$^{\text{Th}}$ & PQ$^{\text{St}}$ & SQ$^{\text{St}}$ & RQ$^{\text{St}}$ \\
    \toprule[0.2em]
    Light Wider MNV2 $^\dagger$ & 17.3 & 66.2 & 22.7 & 9.1 & 62.6 & 12.8 & 28.2 & 71.0 & 35.8 \\
    Light Wider MNV2 & 22.6 & 69.6 & 29.3 & 15.4 & 67.2 & 21.1 & 32.1 & 72.9 & 40.2\\
    Wider MNV2 & 25.3 & 70.6 & 32.3 & 17.6 & 65.5 & 23.4 & 35.5 & 77.4 & 44.0 \\
    Xception-71 & 31.6 & 75.5 & 40.1 & 25.0 & 73.4 & 33.1 & 40.3 & 78.3 & 49.3 \\
    \bottomrule[0.1em]
  \end{tabular}
  }
  \caption{DeeperLab performance on the Mapillary Vistas test set. $^\dagger$: input size is downsampled by 2 ($721\times721$).}
  \label{tab:mapillary_test}
\end{table}

\subsection{Ablation Study}
\label{subsec:ablation_study}

\begin{figure}
     \centering
     \begin{subfigure}[b]{0.2\textwidth}
         \centering
         \includegraphics[width=\textwidth]{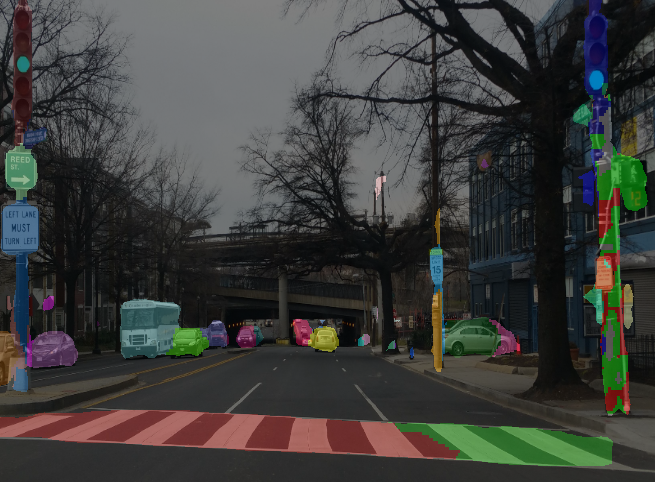}
         \caption{MobileNetV2}
         \label{fig:instance_mnv2}
     \end{subfigure} 
     \hfill
     \begin{subfigure}[b]{0.2\textwidth}
         \centering
         \includegraphics[width=\textwidth]{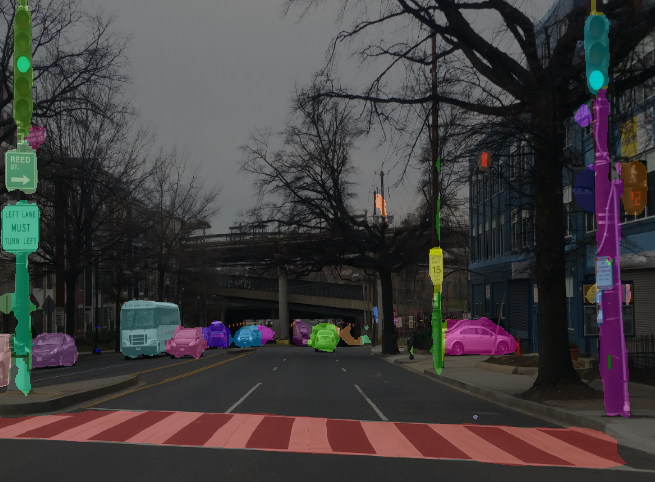}
         \caption{Wider MobileNetV2}
         \label{fig:instance_wmnv2}
     \end{subfigure}
        \caption{Instance segmentation results obtained by directly applying $1\times1$ convolutions to the feature maps extracted by MobileNetV2 and Wider MobileNetV2 (\ie, no ASPP or any other modules). Because of the larger receptive field, Wider MobileNetV2 can segment large objects better (\eg, the crosswalk and the rightmost pole).}
        \label{fig:visual_mnv2_wmnv2}
\end{figure}

{\bf Wider MobileNetV2 backbone design:} The original MobileNetV2~\cite{mobilenetv22018} employs $3\times3$ kernels in all convolutions. We experiment with different kernel sizes, such as $5\times5$ or $7\times7$, to enlarge the network's receptive field. As shown in~\tabref{tab:number_mnv2_wmnv2}, increasing the kernel size is a very effective approach. The PQ and PC are improved by 1.75\% and 4.54\% respectively when the $5\times5$ kernel size is used on Mapillary Vistas, which contains images with much higher resolutions than ImageNet. See \figref{fig:visual_mnv2_wmnv2} for a visual result. We opt for the $5\times5$ kernel size since using the $7\times7$ kernel size only marginally improves the performance, and the resulting network backbone is referred to as Wider MobileNetV2. Additionally, adopting the ASPP module~\cite{chen2017rethinking} further improves accuracy for all settings.

\begin{table}[!t]
  \centering
  \scalebox{0.7}{
  \begin{tabular}{c c | c c}
    \toprule[0.2em]
    Kernel Size & ASPP & PQ (\%) & PC (\%) \\
    \toprule[0.2em]
    $3\times3$ &   & 16.17 & 34.80 \\
    $5\times5$ &   & 17.92 & 39.34 \\
    $7\times7$ &   & 18.27 & 40.33 \\
    \midrule
    $3\times3$ & \checkmark  & 19.21 & 41.07 \\
    $5\times5$ & \checkmark  & 19.85 & 42.98  \\
    $7\times7$ & \checkmark  & 20.14 & 43.40 \\
    \bottomrule[0.1em]
  \end{tabular}
  }
  \caption{The comparison between different Wider MobileNetV2 designs. Employing a larger kernel size in all the convolutions of MobileNetV2 significantly improves the accuracy on Mapillary Vistas, where the images have much larger resolutions than ImageNet. Moreover, the ASPP module is effective for all settings.}
  \label{tab:number_mnv2_wmnv2}
\end{table}

{\bf Decoder and prediction head design:} Given the Wider MobileNetV2 augmented with the ASPP module, we experiment with different decoder architectures in~\tabref{tab:decoder_wider_mobilenet}. The baseline, attaining a performance of 19.85\% PQ and 42.98\% PC, is obtained by directly attaching prediction heads (only one $1\times1$ convolution) to the feature maps. With a simple decoder, which concatenates the bilinearly upsampled (\textbf{BU}) reduced ASPP output (stride = 16) with the reduced low-level feature (stride = 4), the accuracy is improved by 0.93\% PQ and 0.85\% PC. We find that it is effective to increase the number of layers in the prediction heads. Adding one more $3\times3$ convolution layer in all the prediction heads ({\bf DH}) further improves accuracy by 0.81\% PQ and 1.12\% PC. By enlarging the convolution kernel size from $3\times3$ to $7\times7$ in both the decoder and the extra convolution in the heads, the model achieves 22.31\% PQ and 44.62\% PC. Last, the accuracy is significantly improved by replacing the bilinear upsampling strategy by the proposed S2D/D2S strategy, reaching 23.48\% PQ and 46.33\% PC.

\begin{table}[!t]
  \centering
  \scalebox{0.7}{
  \begin{tabular}{c c c c | c c}
    \toprule[0.2em]
    BU & S2D & DH & LK & PQ (\%) & PC (\%) \\
    \toprule[0.2em]
    &            &            &            & 19.85 & 42.98 \\
    \midrule
    \checkmark &            &            &            & 20.78 & 43.83 \\
    \checkmark &            & \checkmark &            & 21.59 & 44.95 \\
    \checkmark &            & \checkmark & \checkmark & 22.31 & 44.62 \\
    \midrule
               & \checkmark &            &            & 21.12 & 44.86 \\
               & \checkmark & \checkmark &            & 22.45 & 46.30 \\
               & \checkmark & \checkmark & \checkmark & 23.48 & 46.33 \\
    \bottomrule[0.1em]
  \end{tabular}
  }
  \caption{The comparison between different decoder and prediction head designs with Wider MobileNetV2 augmented with ASPP. The baseline is obtained by directly attaching the prediction heads (one $1\times1$ convolution) to the output. {\bf BU:} Decoder with Bilinear Upsampling. {\bf S2D:} Decoder with D2S and S2D. {\bf DH:} Deeper Heads with one more extra convolution in the prediction heads. {\bf LK:} Using Larger $7\times7$ Kernels in both the decoder and the extra prediction head.}
  \label{tab:decoder_wider_mobilenet}
\end{table}

{\bf Hard pixel mining:} We find hard pixel mining (\textbf{HPM}) beneficial. As explained in \secref{subsubsec:semantic_segmentation_predicton}, we sort the pixels based on their losses and only backpropagate the top 15\% pixels, the PQ is increased by 0.57\%. If we increase the loss weight of instances smaller than $64\times64$ by $3\times$ (\textbf{SI}), the accuracy is improved by 0.2\% PQ. To maximize the accuracy, we combine these two approaches and achieve 0.92\% PQ and 1\% PC improvement over the baseline. 

\begin{table}[!t]
  \centering
  \scalebox{0.7}{
  \begin{tabular}{c c | c c}
    \toprule[0.2em]
    HPM & SI & PQ (\%) & PC (\%) \\
    \toprule[0.2em]
    &    &  24.07 & 48.23 \\
    \checkmark &            & 24.64 & 48.34 \\
               & \checkmark & 24.27 & 48.42 \\
    \checkmark & \checkmark & 24.99 & 49.23 \\
    \bottomrule[0.1em]
  \end{tabular}
  }
  \caption{The comparison between different hard pixel mining approaches. {\bf HPM:} Hard Pixel Mining. {\bf SI:} Larger loss weights on Small Instances. The accuracy is maximized when both approaches are employed.}
  \label{tab:hard_pixel_mining}
\end{table}

{\bf Directed keypoint relation graph:} We compare two different directed keypoint relation graphs. The first one is the star graph as explained in Sec.~\ref{subsubsec:instance_segmentation_prediction}. Another one is the rectangular graph, where the keypoints are connected in a rectangular shape, and there is no mass-center keypoint. As shown in~\tabref{tab:graph_structure}, using the star graph results in 0.53\% higher PQ and 1.89\% higher PC. We think the mass-center keypoint is important for instance detection.

\begin{table}[!t]
  \centering
  \scalebox{0.7}{
  \begin{tabular}{c c | c c}
    \toprule[0.2em]
    Star & Rectangle & PQ (\%) & PC (\%) \\
    \toprule[0.2em]
    \checkmark &            & 24.07 & 48.23 \\
               & \checkmark & 23.54 & 46.44 \\
    \bottomrule[0.1em]
  \end{tabular}
  }
  \caption{The comparison between alternative directed keypoint relation graphs. Employing the star graph where the mass-center keypoint is connected to the other four keypoints leads to higher accuracy than the rectangle graph.}
  \label{tab:graph_structure}
\end{table}

{\bf Deeper network backbone:} In \tabref{tab:x71}, we report the ablation study with Xception-71~\cite{chollet2016xception,dai2017coco,deeplabv3plus2018} as the network backbone. Our best Xception-71-based model attains an accuracy of 30.46\% PQ and 54.55\% PC on the validation set.

\begin{table}[!t]
  \centering
  \scalebox{0.7}{
  \begin{tabular}{c c c c c | c c}
    \toprule[0.2em]
    ASPP & BU & S2D & HPM & SI & PQ (\%) & PC (\%) \\
    \toprule[0.2em]
    \checkmark & \checkmark &            &            &            & 27.79 & 50.80 \\
    \checkmark &            & \checkmark &            &            & 28.34 & 50.88 \\
    \checkmark &            & \checkmark & \checkmark &            & 30.04 & 53.35 \\
               &            & \checkmark & \checkmark & \checkmark & 30.46 & 54.55 \\
    \bottomrule[0.1em]
  \end{tabular}
  }
  \caption{Employing Xception-71 as the network backbone with different methods. {\bf ASPP:} Encoder with Atrous Spatial Pyramid Pooling. {\bf BU:} Decoder with Bilinear Upsampling. {\bf S2D:} Decoder with D2S and S2D. {\bf HPM:} Hard Pixel Mining. {\bf SI:} Larger loss weights on Small Instances.}
  \label{tab:x71}
\end{table}

%% file: 5_conclusion.tex
\section{Conclusion}
\label{sec:conclusion}

We have proposed and demonstrated the effectiveness of the image parser, DeeperLab, for the challenging whole image parsing task. Our proposed model design attains a good trade-off between accuracy and speed. This is made possible by adopting a single-shot, bottom-up, and single-inference paradigm and integrating various design innovations. These innovations include extensively applying depthwise separable convolution, using a shared decoder output with simple two-layer prediction heads, enlarging kernel sizes instead of making the network deeper, employing space-to-depth and depth-to-space rather than upsampling, and performing hard data mining. Moreover, we have also proposed the `Parsing Covering` (PC) metric to evaluate the parsing accuracy from the region based perspective. We hope the design strategies and the metric will facilitate future research into image parsing.

\paragraph{Acknowledgments}
We thank Peter Kontschieder for the valuable discussion about the Mapillary Vistas result format; Florian Schroff, Hartwig Adam, and Mobile Vision team for support.

%% file: 6_appendix.tex
In this supplementary material, 
\begin{itemize}
    \item we show the results of \emph{DeeperLab} on other datasets, which are Cityscapes~\cite{Cordts2016Cityscapes}, PASCAL VOC 2012~\cite{everingham2014pascal} and COCO~\cite{lin2014microsoft}. The experimental setting is the same as that mentioned in the main paper, unless otherwise stated.
    \item we provide more visualized results for Mapillary Vistas~\cite{Vemulapalli2016Gaussian}.
\end{itemize}

\section{Performance on Cityscapes}

{\bf Experimental setting:} Without using the extra coarse annotations in Cityscapes~\cite{Cordts2016Cityscapes}, the models are trained on the training set ($2,975$ images) and evaluated on the validation set ($500$ images) with the crop size of $721\times721$.

\begin{table}[!t]
  \centering
  \scalebox{0.6}{
  \begin{tabular}{c c | c c c c c}
    \toprule[0.2em]
    Method & Input Size & PQ (\%) & PC (\%) & fps (CPU) & fps (GPU) & Merge (ms)\\
    \toprule[0.2em]
    Light Wider MNV2 & $513\times1025$  & 39.32 & 64.66 & 0.96 & 23.99 & 45 \\
    Light Wider MNV2 & $1025\times2049$ & 48.06 & 69.70 & 0.24 & 10.21 & 154 \\
    Wider MNV2 & $1025\times2049$ & 52.33 & 74.04 & 0.10 & 6.71 & 154\\
    Xception-71 & $1025\times2049$ & 56.53 & 75.63 & 0.07 & 3.24 & 154\\
    \midrule
    Li \etal \cite{li2018weakly} & - & 53.8 & - & - & -\\
    \bottomrule[0.1em]
  \end{tabular}
  }
  \caption{DeeperLab performance on the Cityscapes validation set. Xception-71 based model attains higher accuracy than~\cite{li2018weakly} while Wider MobileNetV2 (Wider MNV2) based model achieves faster inference with comparable accuracy. The model can be further sped up by simplifying the decoder structure (Light Wider MNV2) with a small accuracy drop. With downsampled inputs, Light Wider MNV2 can reach near real-time speed.}
  \label{tab:cityscapes_val}
\end{table}

\begin{figure*}[!t]
    \centering
    \includegraphics[width=0.97\textwidth,keepaspectratio]{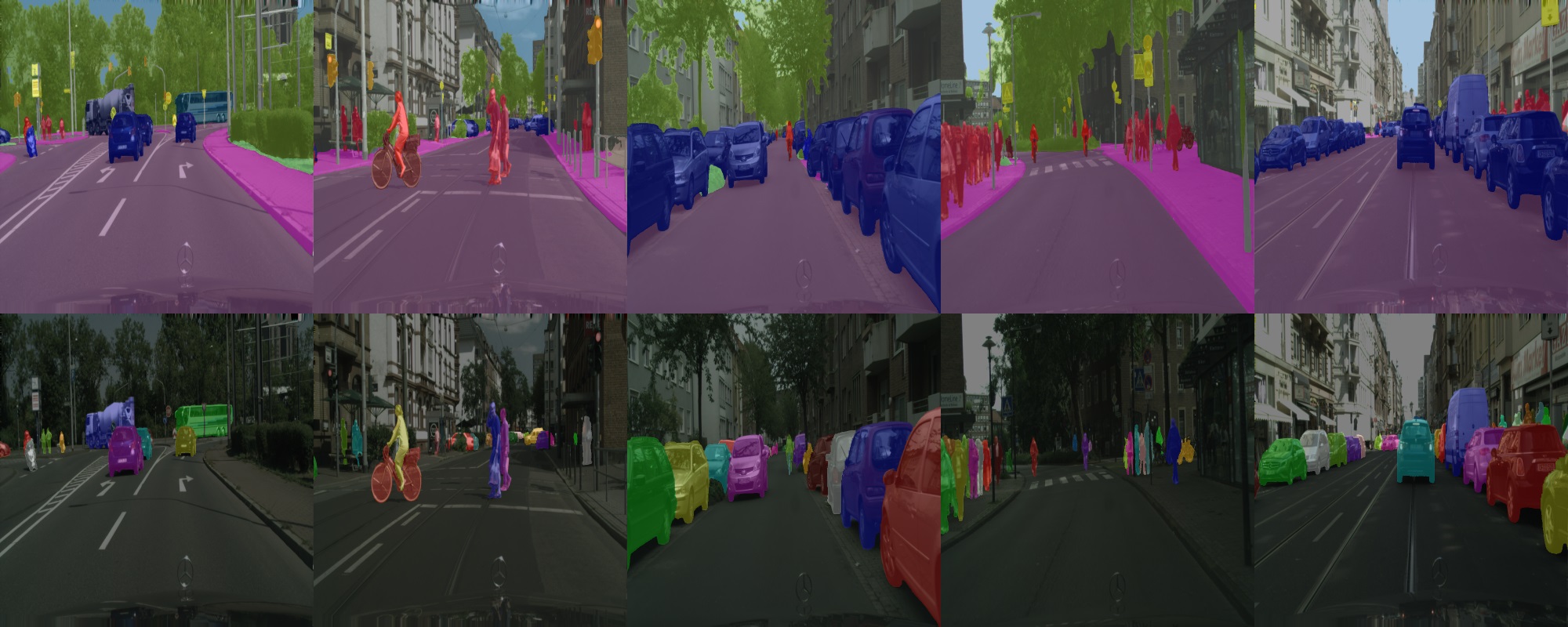}
    \caption{A few image parsing results overlaid on the original images on the Cityscapes validation set with the proposed DeeperLab based on Xception-71. The first row is the predicted semantic segmentation and the second row is the predicted instance segmentation. Note that our model does not generate any VOID labels.}
    \label{fig:xception_visualization_overlay_cityscapes}
\end{figure*}

\begin{figure*}[!t]
    \centering
    \includegraphics[width=0.97\textwidth,keepaspectratio]{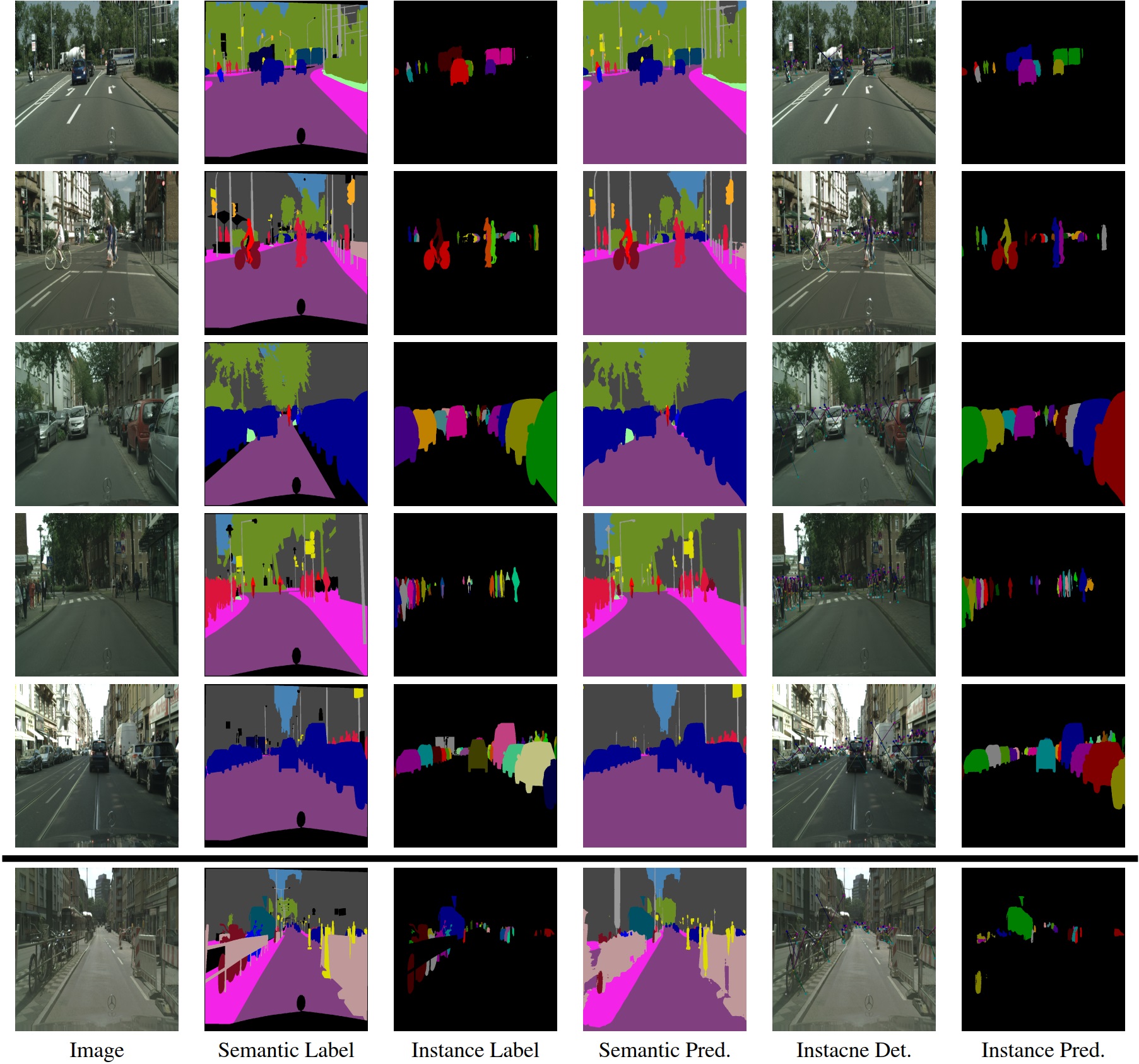}
    \caption{A few image parsing results compared with the ground-truth labels on the Cityscapes validation set with the proposed DeeperLab based on Xception-71. The last row is a failure case. Note that the `Instance Detection` column shows the raw outputs of the keypoints and middle-range offsets of the detected instances, which will be further refined in the later stage.}
    \label{fig:xception_visualization_separate_cityscapes}
\end{figure*}

{\bf Validation set performance:} We summarize the accuracy and speed of DeeperLab on the validation set of Cityscapes in~\tabref{tab:cityscapes_val}. Our Xception-71 based model outperforms~\cite{li2018weakly} in terms of both Panoptic Quality (PQ) and Parsing Covering (PC), and our Wider MobileNetV2 achieves comparable accuracy at the speed of 6.71 fps on GPU. Moreover, our Light Wider MobileNetV2 with downsampled inputs attains near real-time speed (23.99 fps) on GPU. \figref{fig:xception_visualization_overlay_cityscapes} and \figref{fig:xception_visualization_separate_cityscapes} show the qualitative results.

\section{Performance on PASCAL VOC 2012}

{\bf Experimental setting:} We augment the training set of the original PASCAL VOC 2012~\cite{everingham2014pascal} with the extra annotations provided by~\cite{hariharan2011semantic}, resulting in $10,582$ training images (\textit{train\_aug}). The models are trained on this \textit{train\_aug} set and evaluated on the validation set ($1,449$ images).

\begin{table}[!t]
  \centering
  \scalebox{0.6}{
  \begin{tabular}{c c | c c c c c}
    \toprule[0.2em]
    Method & Input Size & PQ (\%) & PC (\%) & fps (CPU) & fps (GPU) & Merge (ms)\\
    \toprule[0.2em]
    Light Wider MNV2 & $257\times257$ & 40.16 & 59.76 & 5.65 & 40.82 & 5 \\
    Light Wider MNV2 & $513\times513$ & 54.09 & 71.09 & 1.83 & 35.01 & 19 \\
    Wider MNV2 & $513\times513$ & 58.75 & 72.72 & 0.77 & 23.76 & 19 \\
    Xception-71 & $513\times513$ & 67.35 & 77.57 & 0.57 & 13.93 & 19 \\
    \midrule
    Li \etal \cite{li2018weakly} $^\dagger$ & - & 62.7 & - & - & - & - \\
    Li \etal \cite{li2018weakly} $^\ddagger$ & - & 63.1 & - & - & - & - \\
    \bottomrule[0.1em]
  \end{tabular}
  }
  \caption{DeeperLab performance on the PASCAL VOC 2012 validation set. Xception-71 based model attains higher accuracy than~\cite{li2018weakly} without pretraining on COCO~\cite{lin2014microsoft}. Moreover, the simplified Wider MNV2 (Light Wider MNV2) reaches real-time speed without downsampling inputs. $^\dagger$: fully-supervised without COCO. $^\ddagger$: fully-supervised with COCO.}
  \label{tab:pascal_val}
\end{table}

\begin{figure*}[!t]
    \centering
    \includegraphics[width=0.97\textwidth,keepaspectratio]{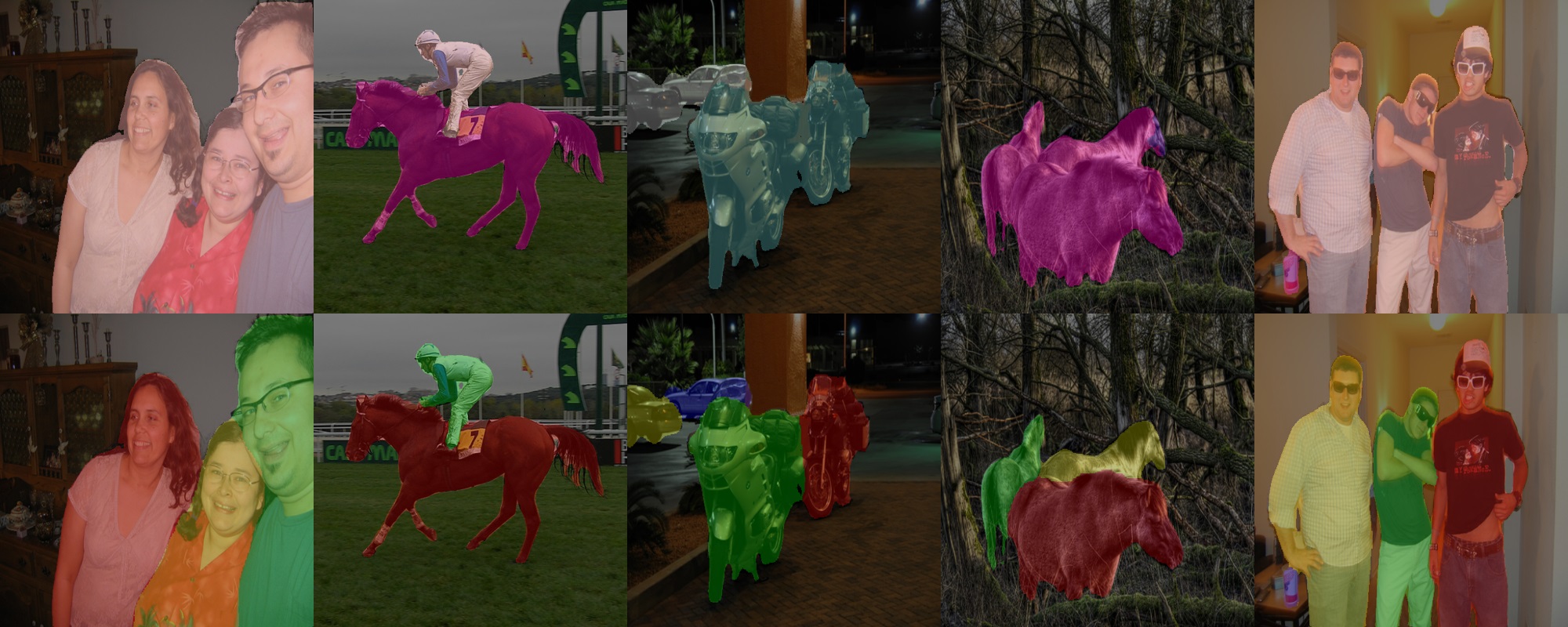}
    \caption{A few image parsing results overlaid on the original images on the Pascal VOC 2012 validation set with the proposed DeeperLab based on Xception-71. The first row is the predicted semantic segmentation and the second row is the predicted instance segmentation. Note that our model does not generate any VOID labels.}
    \label{fig:xception_visualization_overlay_pascal}
\end{figure*}

\begin{figure*}[!t]
    \centering
    \includegraphics[width=0.97\textwidth,keepaspectratio]{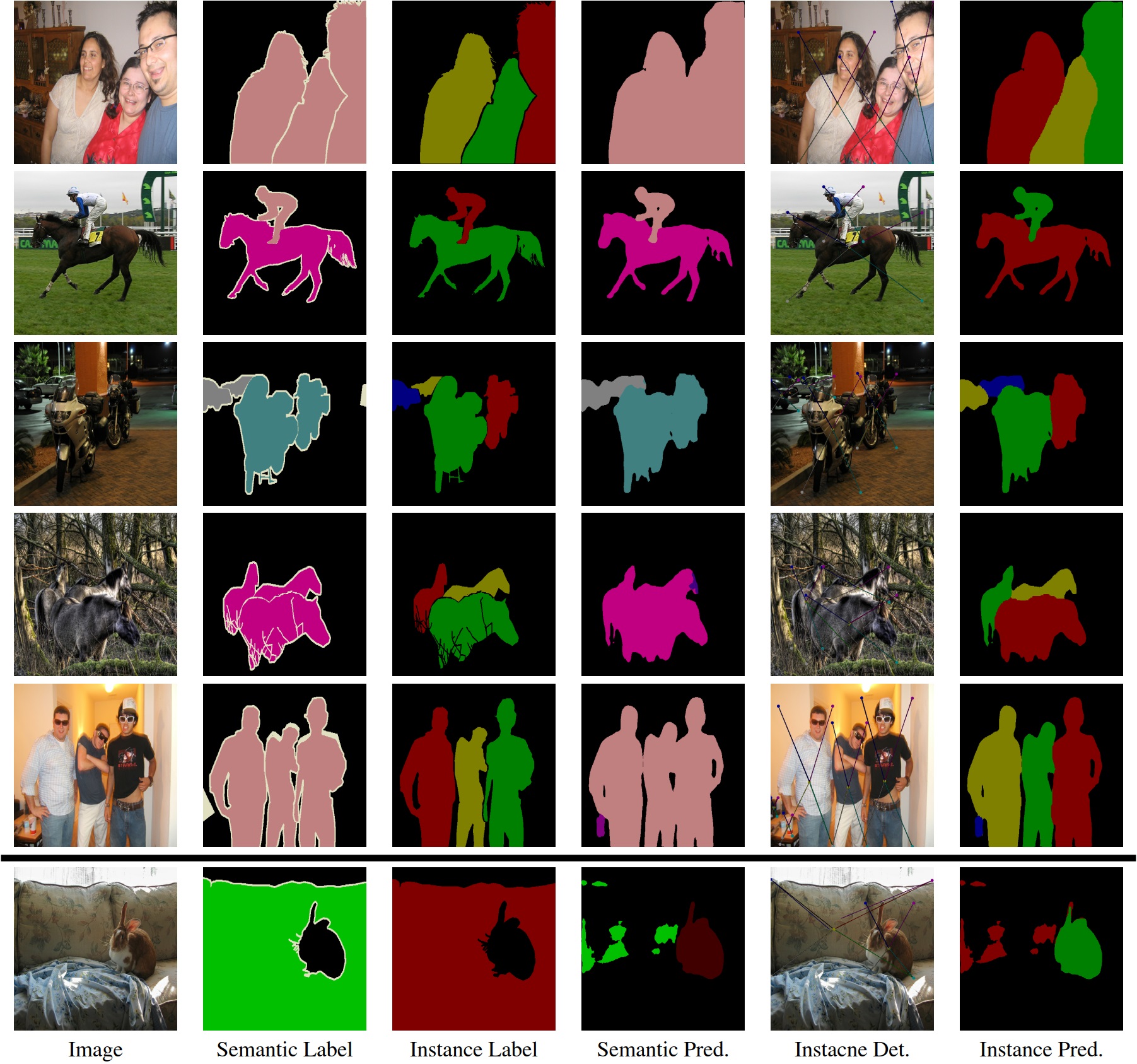}
    \caption{A few image parsing results compared with the ground-truth labels on the Pascal VOC 2012 validation set with the proposed DeeperLab based on Xception-71. The last row is a failure case. Note that the `Instance Detection` column shows the raw outputs of the keypoints and middle-range offsets of the detected instances, which will be further refined in the later stage.}
  \label{fig:xception_visualization_separate_pascal}
\end{figure*}

{\bf Validation set performancet:} We summarize the accuracy and speed of DeeperLab on the validation set of PASCAL VOC 2012 in~\tabref{tab:pascal_val}. Our Xception-71 based model outperforms~\cite{li2018weakly} in terms of both Panoptic Quality (PQ) and Parsing Covering (PC) even without pretraining on COCO~\cite{lin2014microsoft}. Moreover, our Light Wider MobileNetV2 attains real-time speed (35.01 fps on GPU) without downsampling inputs. \figref{fig:xception_visualization_overlay_pascal} and \figref{fig:xception_visualization_separate_pascal} show the qualitative results.

\section{Performance on COCO}

{\bf Experimental setting:} Although enlarging images has been shown to be effective on COCO~\cite{lin2016feature,he2017mask}, we do not upsample the input images because of the consideration of speed. We leave exploring this augmentation as future work and focus on high-speed single-shot models in this work. Moreover, we do not perform hard pixel mining on COCO because it hurts the accuracy.

\begin{table}[!t]
  \centering
  \scalebox{0.6}{
  \begin{tabular}{c c | c c c c c}
    \toprule[0.2em]
    Method & Input Size & PQ (\%) & PC (\%) & fps (CPU) & fps (GPU) & Merge (ms)\\
    \toprule[0.2em]
    Light Wider MNV2 & $321\times321$ & 17.51 & 39.15 & 3.13 & 33.84 & 8 \\
    Light Wider MNV2 & $641\times641$ & 24.10 & 48.47 & 0.88 & 20.81  & 25 \\
    Wider MNV2 & $641\times641$ & 27.91 & 52.38 & 0.43 & 17.19 & 25 \\
    Xception-71 & $641\times641$ & 33.79 & 56.82 & 0.33 & 10.59 & 25 \\
    \bottomrule[0.1em]
  \end{tabular}
  }
  \caption{DeeperLab performance on the COCO validation set. Xception-71 based model attains higher accuracy while Wider MobileNetV2 (Wider MNV2) based model achieves faster inference. The model can be sped up by simplifying the decoder structure (Light Wider MNV2) with a small accuracy drop. With downsampled inputs, Light Wider MNV2 can reach real-time speed. Note our Xception-71 based model attains the mIOU of 55.26\% for semantic segmentation task.}
  \label{tab:coco_val}
\end{table}

\begin{figure*}[!t]
    \centering
    \includegraphics[width=0.97\textwidth,keepaspectratio]{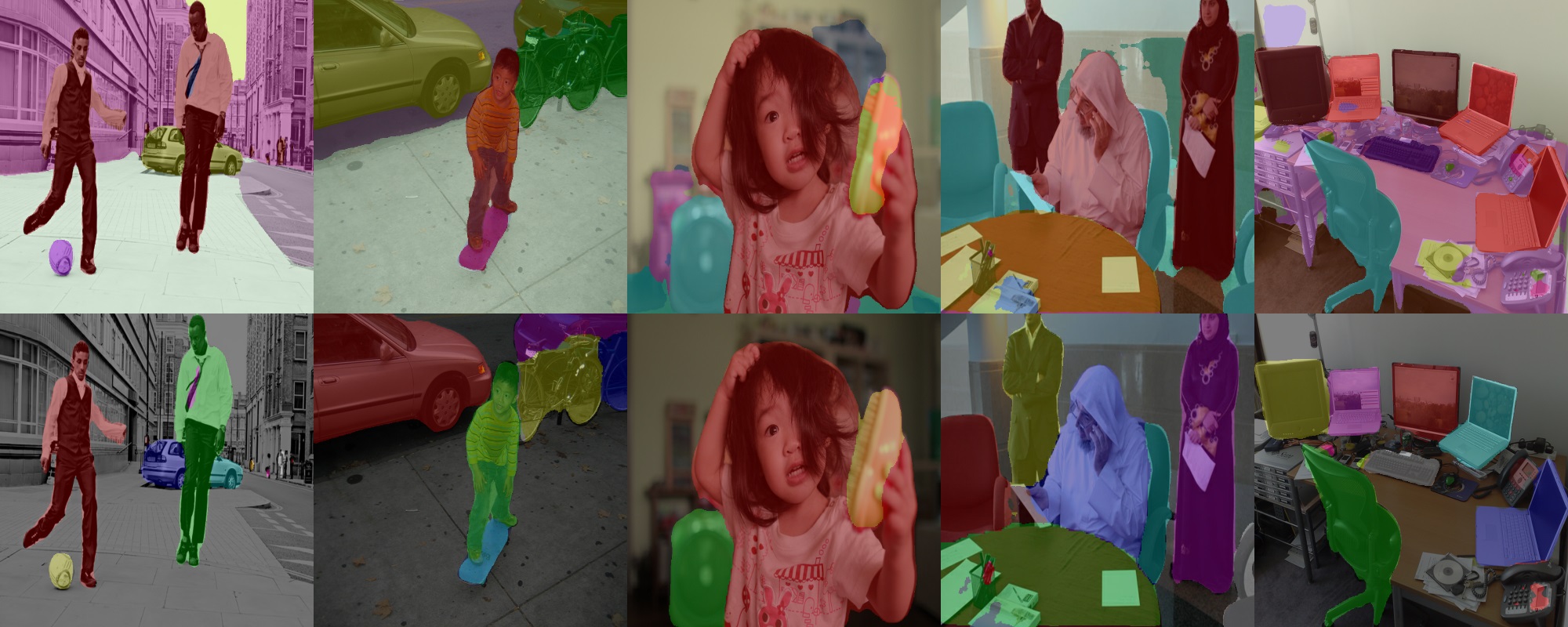}
    \caption{A few image parsing results overlaid on the original images on the COCO validation set with the proposed DeeperLab based on Xception-71. The first row is the predicted semantic segmentation and the second row is the predicted instance segmentation. Note that our model does not generate any VOID labels.}
    \label{fig:xception_visualization_overlay_coco}
\end{figure*}

\begin{figure*}[!t]
    \centering
    \includegraphics[width=0.97\textwidth,keepaspectratio]{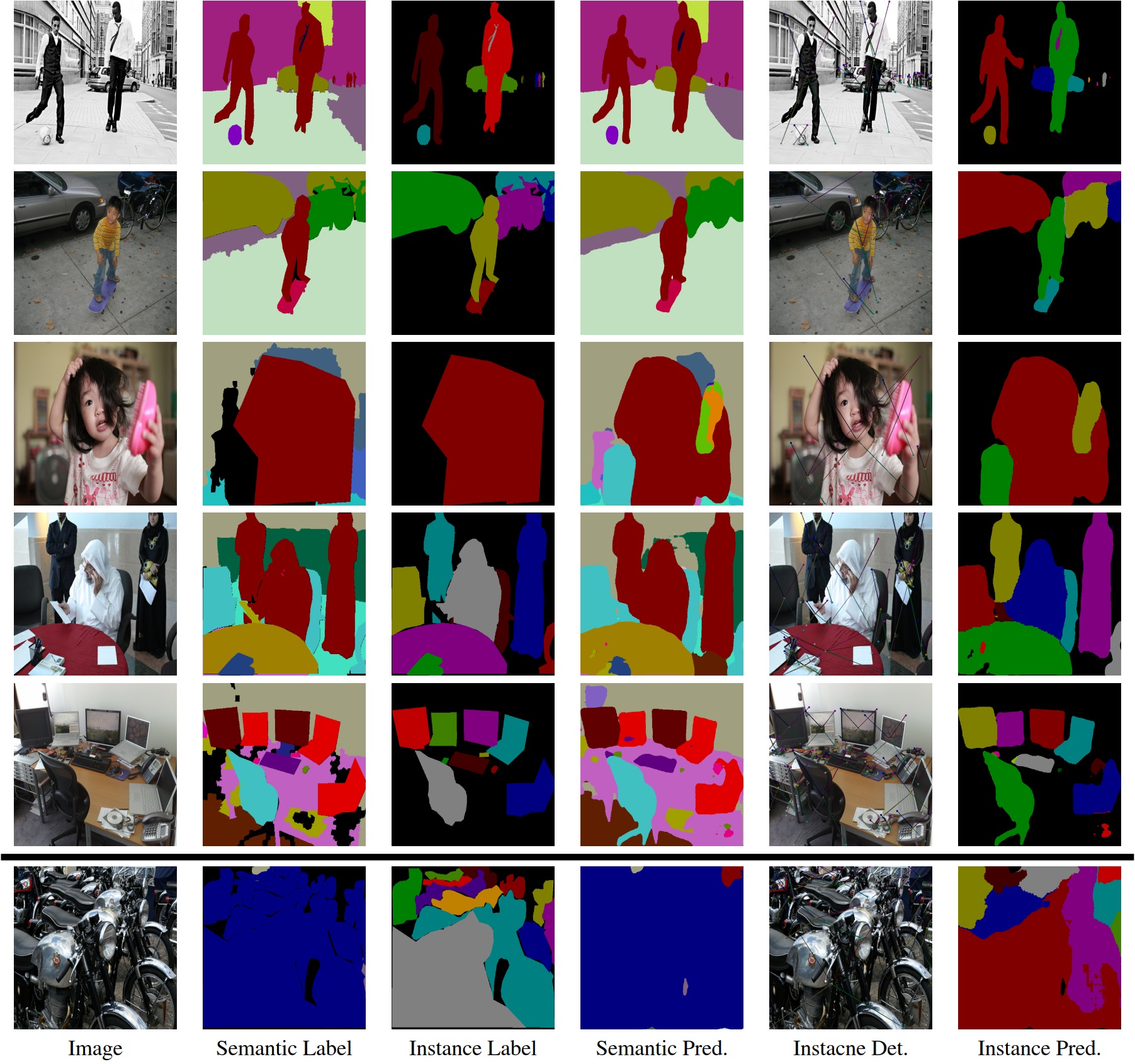}
    \caption{A few image parsing results compared with the ground-truth labels on the COCO validation set with the proposed DeeperLab based on Xception-71. The last row is a failure case. Note that the `Instance Detection` column shows the raw outputs of the keypoints and middle-range offsets of the detected instances, which will be further refined in the later stage.}
  \label{fig:xception_visualization_separate_coco}
\end{figure*}

{\bf Validation set performance:} We summarize the accuracy and speed of DeeperLab on the validation set of COCO in~\tabref{tab:coco_val}. Our Xception-71 based model attains 33.79\% Panoptic Quality (PQ) and 56.82\% Parsing Covering (PC) at the speed of 10.59 fps on GPU. Wider MobileNetV2 based model increases the speed to 17.19 fps on GPU at the cost of accuracy. Our Light Wider MobileNetV2 with downsampled inputs further pushes the speed to 33.84 fps on GPU. \figref{fig:xception_visualization_overlay_coco} and \figref{fig:xception_visualization_separate_coco} show the qualitative results.

\begin{table}[!t]
  \centering
  \scalebox{0.61}{
  \begin{tabular}{c | c c c | c c c | c c c}
    \toprule[0.2em]
    Method & PQ & SQ & RQ & PQ$^{\text{Th}}$ & SQ$^{\text{Th}}$ & RQ$^{\text{Th}}$ & PQ$^{\text{St}}$ & SQ$^{\text{St}}$ & RQ$^{\text{St}}$ \\
    \toprule[0.2em]
    Light Wider MNV2 $^\dagger$ & 18.0 & 71.3 & 23.6 & 18.5 & 71.8 & 24.3 & 17.2 & 70.5 & 22.6 \\
    Light Wider MNV2 & 24.5 & 73.2 & 31.5 & 26.9 & 73.7 & 34.6 & 20.9 & 72.5 & 26.9\\
    Wider MNV2 & 28.1 & 75.3 & 35.8 & 30.8 & 75.7 & 39.1 & 24.1 & 74.6 & 30.9 \\
    Xception-71 & 34.3 & 77.1 & 43.1 & 37.5 & 77.5 & 46.8 & 29.6 & 76.4 & 37.4 \\
    \bottomrule[0.1em]
  \end{tabular}
  }
  \caption{DeeperLab performance on the COCO test-dev set. The numbers on the test-dev set are very close to that on the validation set (\tabref{tab:coco_val}). $^\dagger$: input size is downsampled by 2 ($321\times321$).}
  \label{tab:coco_testdev}
\end{table}

{\bf Test-dev set performance:} The performance of our models on the test-dev set of COCO is reported in~\tabref{tab:coco_testdev}. We can see that the numbers on the test-dev set are very close to that on the validation set (\tabref{tab:coco_val}).

\section{Performance on Mapillary Vistas}

\begin{figure*}[!t]
    \centering
    \includegraphics[width=0.97\textwidth,keepaspectratio]{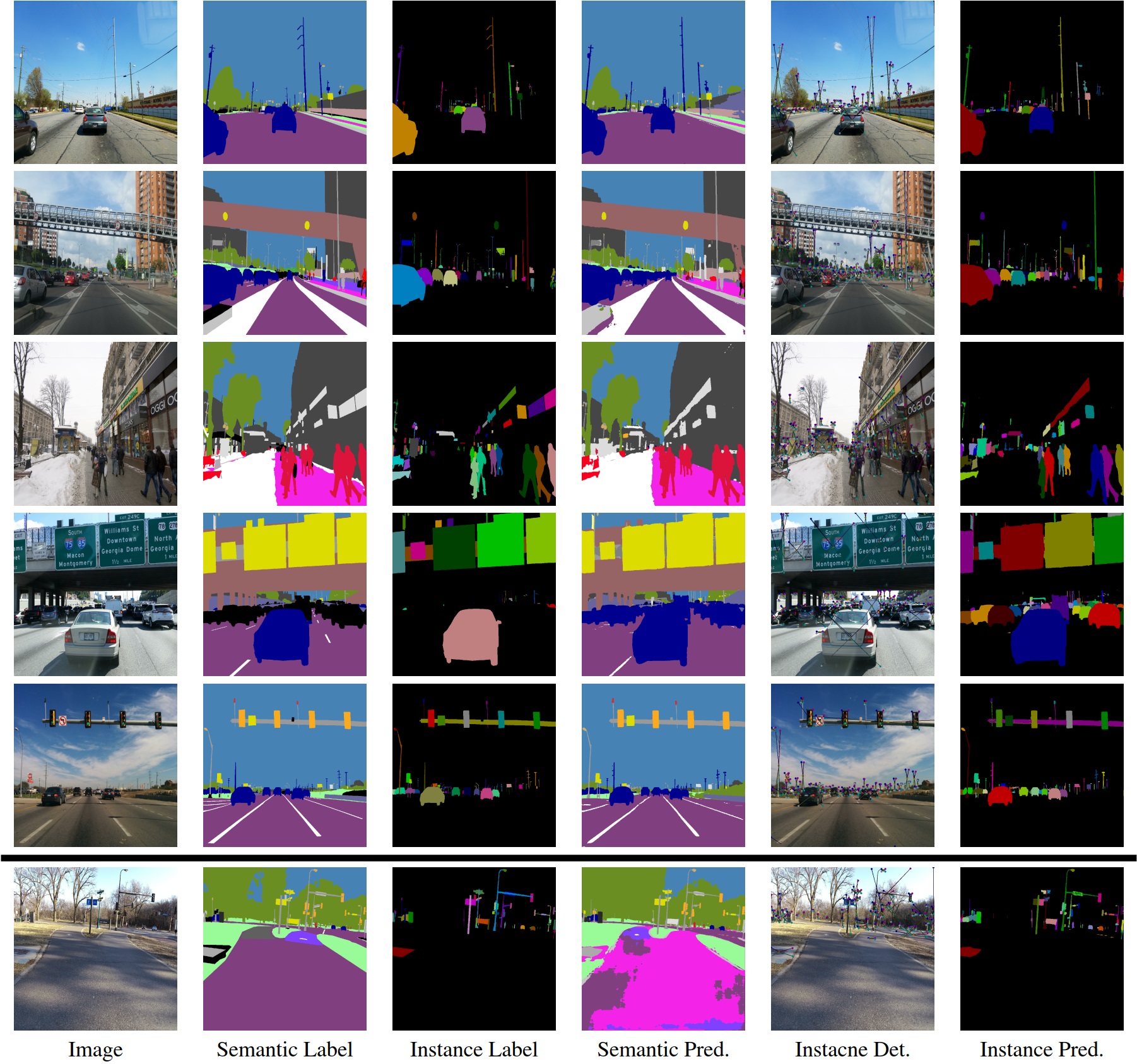}
    \caption{A few image parsing results compared with the ground-truth labels on the Mapillary Vistas validation set with the proposed DeeperLab based on Xception-71. The last row is a failure case. Note that the `Instance Detection` column shows the raw outputs of the keypoints and middle-range offsets of the detected instances, which will be further refined in the later stage.}
  \label{fig:xception_visualization_separate_mapillary}
\end{figure*}

\figref{fig:xception_visualization_separate_mapillary} shows the extra qualitative results.

\clearpage

